\documentclass[11pt]{article}
\usepackage[a4paper, hcentering=true, vcentering=true, total={15cm, 20.5cm}]{geometry}
\usepackage{expex}
\newcommand{\xref}[1]{~(\ref{#1})}   
\newcommand{\lbr}{\ensuremath{[}}
\newcommand{\rbr}{\ensuremath{]}}
\newcommand{\xreff}[2]{~(\ref{#1}#2)}
\lingset{
    belowglpreambleskip=-0.2ex,
    everygla=,
    aboveglftskip=-0.2ex,
    interpartskip=0ex,
    glspace=!0pt plus .2em,
    glrightskip=0pt plus .5\hsize,
    aboveexskip= .99ex plus .6ex minus .4ex,
    belowexskip=.99ex plus .6ex minus .4ex,
    labelalign=center,                     
    exnotype=arabic,   
    numoffset=0pt,  
    textoffset=.4ex,  
    labeloffset=2pt
}
\usepackage{setspace}

\usepackage{mathptmx}
\usepackage{natbib}
\usepackage{titling,graphicx}
\usepackage{qtree}
\usepackage{ccg-latex-expex2}
\newcommand{\ml}[2]{\multicolumn{#1}{l}{#2}}

\newcommand{\piin}[2]{\ensuremath{#1#2}}      
\newcommand{\piou}[2]{\ensuremath{\bar{#1}#2}}  
\newcommand{\pisi}{\ensuremath{\tau}}           
\newcommand{\pior}{\mbox{\texttt{|}}}  
\newcommand{\pisu}[2]{\ensuremath{#1 + #2}}     
\newcommand{\pina}[2]{\ensuremath{\nu#1#2}}     
\newcommand{\pipa}[2]{\ensuremath{#1 \pior #2}} 
\newcommand{\pire}[1]{\ensuremath{!#1}}         
\newcommand{\pize}{\ensuremath{\mathbf{0}}}              
\newcommand{\plancat}{\ensuremath{\doteq}\,}

\begin{document}
\title{Mobile Sequencers}
\author{Cem Boz\c{s}ahin\\
Department of Cognitive Science\\
Middle East Technical University, Ankara, T\"urkiye}
\date{\texttt{bozsahin@metu.edu.tr}~~May 2024\\[2ex]
\textit{\scriptsize
\hfill  We can only know in the nervous system\\ \hfill what we have known in behavior first.~~\emph{Julian Jaynes 1976}\\[1ex]
\hfill Computer science is no more about computers\\
\hfill than astronomy is about telescopes.~~\emph{Edsger Dijkstra 1990}\\[1ex]
\hfill Psychologists have been concerned chiefly with the question of\\ \hfill whether or not the organizing processes
displayed in serial action\\ \hfill are conscious,
and very little with the organization itself.~~\emph{Karl \citealt{Lash:51}}\\[1ex]
\hfill Religion, it has been said, is in its origin a technique for success. \emph{Alan H \citealt{brod:48}}}
}
\maketitle

\begin{titlingpage}

\begin{abstract}
\noindent The article is an attempt to contribute to  explorations of a common origin for language and planned-collaborative action. It gives `semantics of change' the central stage in the synthesis, from its history and recordkeeping to its development, its syntax, delivery and reception, including substratal aspects. 

It is suggested that to arrive at a common core, linguistic semantics must be understood as studying through syntax mobile agent's representing, tracking and coping with change and no change. Semantics of actions can be conceived the same way, but through plans instead of syntax. The key point  is the following: Sequencing itself, of words and action sequences, brings in more structural interpretation to the sequence  than which is immediately evident from the sequents themselves. Mobile sequencers can be understood as subjects  structuring reporting, understanding and  keeping track of change and no change. The idea invites rethinking of the notion of category, both in language and in planning.

Linguist's search for explaining the gaps in possible structures, and offlineness of language, and computer scientist's search for possible plan landscape, and onlineness of action, are leveraged by the synthesis for open exploration. It leaves very little room for analogies and instrumental thinking, such as language being an infinite gift, or
computer being the ultimate human tool. Nothing is infinite if modern physics is right, not even the computer's name-recursive representations, which is commonly---and misleadingly---compared
with human's value-recursive representations. This has implications for the synthesis.

Understanding understanding change by mobile agents is suggested to be
about human extended practice, not extended-human practice. That's why linguistics is as important as computer science in the synthesis. It must rely on representational history of acts, thoughts and expressions, personal and public, crosscutting overtness and covertness of these phenomena.
It has implication for anthropology in the extended practice, which is covered briefly.
\end{abstract}\newpage
\tableofcontents
\end{titlingpage}

\section{Introduction: language, action and  extended notion of category}
Sequencing of actions and speaking-signing by a human agent bring in more semantics to  the sequence in planning and language. It reveals
more dynamics of apparently serial or serializeable speech/sign-events and action-events unfolding in time and space, which may not be evident from
the individual meanings of the sequents themselves. 

For example, consider buying a ticket, going to the airport,  then flying; versus flying, then buying a ticket. In the first case, the ticket may or may not be for the flight (for example, it may be for the shuttle to the airport). In the second case, the ticket is more likely not  for the flight unless states of affairs suggest otherwise, for example, not being able to pay in full but being allowed on board the flight, to complete the payment later. This unusual connection would demand another interpretation of  events mainly because of its unusual ordering of elements bearing the same meaning as those in normal course of events. 

Now consider the Chinese expression \emph{W\v{o}  y\=ou le Zh\=angs\=an y\'{\i} m\`o} (I secluded-serene \textsc{ASPECT} Zhangsan one silent), meaning
`I teased Zhangsan'.  The meaning of tease depends only on \emph{y\=ou} and \emph{m\`o}, but it rises above independent meanings of \emph{y\=ou} and \emph{m\`o}, respectively `secluded/serene' and `silent', words which cannot be adjacent in expressing this meaning:
*\emph{W\v{o}  y\=ou (le) (yi) m\`o  Zh\=angs\=an}, and *\emph{W\v{o}  y\=ou m\`o (le) Zh\=angs\=an}. 
When adjacent, they constitute an adjective or intransitive verb
(to be humorous), but they do not deliver `teasing someone'. The situation is not too far from the airport example in terms of their structure.
Whatever constitutes the meaning of `tease', there are reasons to believe that it depends on the elements combined, also in what order and how we combine them, rather than combining the parts meaning seclude/serene and silent, then post-interpreting them as tease. This is because  there are
syntactic constraints in the combination if the narrow meaning (tease) is to be maintained, and freer syntax if it is not. 

We can see the difference of syntactic constraints
in two English idioms with varying syntactic behavior, spill the beans (for `divulge') and kick the bucket (for `die'). Compare their syntactic flexibility: \emph{spilling the beans is fashionable} and \emph{the beans that you spilled are juicy}, versus \emph{kicking the bucket while on vacation} and
\emph{the bucket that you kicked saddened}, where in the latter the idiomatic meaning is not available in freer syntax. From syntactic freedom it does not follow that the meaning
of an expression is just as free; it depends on how the two, syntax and semantics, are connected. From syntactic narrowness or inflexibility it does not follow that 
we are looking at meaning assembly dissociated from grammar, for example the meaning of kicking the bucket. We need finer study of linguistic categories to see that both ways of thinking are oversimplifications. 

Another example from Chinese may clarify interaction of syntactic and semantic decomposability. In all three examples of\xref{eg:intro}, the narrow meaning of `becoming angry' is the only one available \citep{kao:24}, arising from `generate' and `air'. Yet they show in
(a) potential lexical atomicity of `generate' and `air', in (b) syntactic adjunction between them, and in (c) resultative complementation between them. We have to explain how meaning
can stay narrow  when syntax decomposes, in the process also revealing independent meanings of \emph{sh\=eng} and \emph{q\`{\i}}. 

\pex\label{eg:intro}
 \a\begingl
 \gla Zh\=angs\=an  \underline{sh\=eng} \underline{q\`{\i}} le//
 \glb Zhangsan generate air \textsc{asp}//
 \glft  `Zhangsan got angry.' \trailingcitation{\citealt{kao:24}:1}//
\endgl
 \a\begingl
 \gla Zh\=angs\=an  \underline{sh\=eng}  le  h\v{a}od\`{a}  de \underline{q\`{\i}}//
 \glb Zhangsan generate  \textsc{asp}  huge \textsc{nom}  air//
 \glft `Zhangsan got very angry.' (lit. `Zhangsan generated huge air.')//
 \endgl
 \a\begingl
 \gla Zh\=angs\=an \underline{sh\=eng}  w\'an \underline{q\`{\i}} le//
 \glb Zhangsan generate  finish  air  \textsc{asp}//
 \glft `Zhangsan stop being angry.' (lit. `Zhangsan finished generating air.')//
 \endgl
\xe

In \cite{bozs:22-jlli}, I proposed that such properties are not outliers of a linguistic system or signs of a parallel/auxiliary system for idiosyncrasy (schemas, metaphors etc.). They reveal the categorial aspect of combining meanings \emph{in general}. They pose a challenge to current understanding of categories-as-labels idea that is common in linguistics: Syntactic decomposability does not entail semantic decomposability, therefore we have to explain what keeps the narrow meaning intact. Leaving the problem to auxiliary mechanisms sweeps it under the carpet.

The notion of complex category that categorial grammarians employ ever since \cite{huss:00,lesn:29,ajdu:35,Mont:73} seems to be necessary and sufficient for lexically controlling decomposability on both sides, that is, in language and planning.
Such categories are complex and procedural, \emph{building} abstract structure and surface structure of syntax and semantics from sequencing, rather
than post-realizing sequencing from an abstract structure where finer control of syntax and semantics would be left unstudied.

It will be argued in the current article that,
somewhat surprisingly, the semantics extending to both planning and language must be mobile as well as being complex and procedural, full of side-effects. Complex categories of categorial grammarians can be adopted for this purpose; however, they must be extended for a synthesis concentrating on change. 
It must face the Frame Problem, which is understanding, tracking and coping with change and no change,
representation of which must facilitate joint attention and consequences to lack of it. For example, when we fly, the color of the airport walls would not change, but our location would.
We are here talking about abstract representation,
not necessarily computational representation (i.e. we do not have to be specific about the encoding and decoding relations, which we must do in all computations), and obviously not about representations which are claimed to be mental. We are concerned with the representational support that will make analyses hold, because, as we shall see, both language and organized collaborative action are representation-hungry problems.

\section{Synthesis}
It might appear that we are making  language and action, which are complex problems on their own, even more complex by studying them together. However, focusing on understanding the nature of the common problem rather than entangling ourselves 
in tying up their independently worked out solutions may give us a new understanding. 
In Marr (\citeyear{marr:77}) terminology, which we cover shortly, there may be a type-1 theory to study the nature of the common problem,  connecting why, what, how and where, for  common representational core to understand language and planning.

We have some results to reflect on before we start.
On the language side, in linguistics, there is a mathematical result showing the limit to permutation patterns of serial elements to be altogether meaningful by themselves hierarchically. In other words, only some hierarchical structures are locally combinable, therefore not all hierarchies are in the same
boat.  
There is  an independently developed linguistic theory which derives all and only such patterns of syntax-semantics for linguistic expressions.
On the computer side, in computer science, in the manner of Marr \citeyear{marr:77}-style exploration of complex problem solving, there is accumulation of knowledge
in planning community about possible planned action. There are recent attempts
to representationally as well as inferentially support  early studies  of dealing with change and no change, for example
circumscription \citep{McCa:69}. For both developments the Frame Problem of mobile semantics is relevant to understanding because as we interpret the hierarchy the world changes.

It is suggested here that language and planning meet at mustering as much mobile meanings from
meaningful sequences of words and actions as possible so that we can cope with change, anticipate it, and control it. Mobile meanings are meanings owned and acted upon by mobile agents. I will be more explicit as we proceed.

Therefore mobility must be built-in to language as well as daily action and organized behavior, but distinctly represented, for reasons that we shall come to subsequently. Briefly, we must distinguish offlineness of language, that is, letting go of conditions of words although we know what their real world consequences are, and non-offlineability of action.\footnote{This is different than some linguists also arguing for offlineness of language, such as \cite{bickerton96}, who claims no connection to action in syntax, following the commonly held Chomskyan position. However, the common progenitor they cite for this view, \citealt{lenn:67}:331, makes it clear that 
\emph{differentiation capacity} is species-specific; he does not claim it to be task- or domain-specific.
It will be seen that it also differs from Mobotics---mobile real world robotics---emanating from \cite{Brooks:91} in not taking the world as its best representation, by attempting to study how we can rise above what we are exposed to.
In the terminology of \cite{clar:94}, language and collaborative planning are representation-hungry problems.} 

Language and action are claimed to differ in degree of offlineability of action, where in language we must let go of change implicated by the expression otherwise we cannot speak or sign and act
asynchronously, and in daily action we cannot let go of change and expect to survive in many circumstances.

I discuss  the historical marks for both domains in the substrate (mind/brain) based on abstract representations.
Their emergence may be due to an evolutionary process that in the end  has probably delivered transparency of speech-plans to speaker's  self  without compromising their potential opaqueness to the hearer, and overt
transparency but teleological opaqueness of plans and actions to participants and interlocutors.
Deceit and opaqueness have survival value in both domains; they can be recipes for success \citep{jayn:76}.
Mobile semantics for meaningful sequences is offered to provide representational and inferential  support for them without having to be specific about their intent or content.

\section{A Marrian question: Why do we express thought?}\label{sec:thought}
Many aspects of human high-level cognitive processes such as those involved in language, planned action, vision, thinking, navigation, collaboration and inference are massively serial, although their centralized substrate, the neural wetware,
affords massive parallelism at the micro level \citep{Rume:86a,koss:88,houg:95,thor:07,quin:13,Fitch:14}. There is probably an  evolutionary explanation for this gap \citep{Lash:51,terr:94,mcgo:96,tarr:94,fitc:08}.
Computation's role in the explanation must be studied with precaution otherwise it runs the risk of pretending to be panacea for all our problems. 

It is possible that planning and collaborative action may reach a type-1 joint theory  addressing all aspects of  Marr's (\citeyear{marr:77}) trilevel formulation of understanding  human problems via computation. 

He suggested that level-1 is the  problem level, the level at which we wish to formulate a theory to understand the nature of the problem, level-2 is the solution (or algorithm) level for the problem, and level-3 is the realization (or substrate) level. He considered the top level to be about goals and constraints of a problem domain, primarily addressing `why' questions related to `what' questions, the middle level to be about algorithms and representations, primarily addressing 'how' and `what', and the bottom level to be about the substrate, primarily addressing `where', `who' and `in what way', for example
where is the mind, and who does it how. The levels primarily address different concerns. Level-1 is about understanding a problem, level-2 is about understanding the solutions to the problem, and level-3 is about understanding the realizations of the problem and its solutions. 

According to Marr, type-1 theories manage to put these levels together by decomposition, whereas type-2 theories fall short of that in some respect.
In a way, we know a type-1 theory when we see one. We would probably feel the same way about type-2 theories, but for a different reason. They do not decompose like  type-1 theories to reveal the distinction between the nature of the problem and the nature of the solutions to the problem. Their  interaction altogether  would be their simplest description.

He also  proposed
that a desire to change a more abstract (i.e. lower-numbered, equivalently, higher) level need not be motivated by  the properties of a more concrete (i.e. higher-numbered) level
because we need to have bridge laws anyway across levels.
This need arises from the hope of achieving some explanation of the problem  rather than its description, if the problem lends itself to careful scrutiny at all levels by being explicit about the bridges across levels. 

In this way of thinking 
making the levels distinction would not make us dualists, 
and staying at one level does not take us closer to monists as \cite{dama:12} appears to suggest. The levels distinction attempts to clarify the terms of the proposed explanation and the questions asked in a complex problem.
Fitch's (\citeyear{Fitch:14}) proposed extension of the Marr paradigm
is one more example which seems to be motivated by a stronger demand from a theory of natural  language. 

It is not a coincidence that computer science---which has had a nowadays waning impact in cognitive science except in performance---makes a living precisely from distinguishing understanding of a problem itself and 
understanding of a solution to the problem, as witnessed for example in its complexity theory for
understanding resource use in behavior, its domain theory for understanding representation, its
programming language theory for understanding interpretation, and architecture theory for understanding
execution. And, even for prominent computer scientists who managed to stay out of the mainstream cognitive science boom, such as Dijkstra (see the quote attributed to him in the prologue); 
\cite{knut:96,vali:13}, computer science is more about humans and their practices than the computer itself.

There is a further distinction made by \cite{bozs:turtles} for understanding the differing role of computation at each level. The top level computational concern is the exploration of principles in computer science terms, the middle level computational concern is the mechanism (i.e. abstract computation and representation---the algorithm), and the lowest
level computational concern is about physical constraints where  computation can be an instrumental concept,  rather than constitutive as in the upper levels, because
any constraining function reduces to its calculation at this level anyway \citep{cock:12}.\footnote{Every level demands from computation to target different questions, even different `how' questions. What Marr appears to have meant by `computational theory' as top level is that it is the only noninstrumental level that focuses on the nature of the problem rather than solutions to the problem.}

It seems that linguistics has a level-2+level-3 theory
\citep{stee:20,stan:23}, to explore a type-1 theory of language \emph{and} organized action, although Marr himself
did not consider linguistics to be a likely source for a type-1 theory. The theory solves the problem of identifying configurational limits on expressing human thought via language.

The type-1 theory can attempt  the following question: Why do we express thought? Although this question is at the back of our minds in whatever we do,
it has been difficult to study it until we  discovered its configurational limits in language, which are found to be necessary but insufficient to understand  its effects in environment control by subjects. Environment control is not only relevant to talking and hearing about actions but also to actions themselves and their consequences. It seems like we  now have the tools to study them together.

I make the suggestion now because we have come to a better understanding of (i) the role of local computing in linguistics
and (ii) the pitfalls of the brain-is-a-computer metaphor, which licensed the leap in the minds of many from computational representation to mental representation. In fact, the warnings about the metaphors  have been around for a long time
in computer science \citep{weiz:76}, linguistics \citep{sand:98}, and neuroscience \citep{cise:99}.

I suggest that it is difficult to proceed towards a type-1 candidate theory if we do not look at the level-2+level-3 theories in sufficient detail. I offer one for critique. Like many such theories, it will start at the middle level, assuming the lowest level's physical realizability is relatively well understood.

In return for starting at the middle level, we need to gain on the joint problem (i.e. level-1) of speech-sign and action. I will argue that organized action and linguistic expression of thought primarily differ in offlineability of speech-sign act.  

Offlineness is the property of letting go of conditions of semantics of an expression without losing touch with its meaning. For example, \emph{I walked the dog} reports  a state of affairs involving an agent and patient, but it did not itself realize the event implicated by these affairs, although meaningful continuations can be conceived to show that letting go of the conditions does not imply disregarding the exchanged states of affairs.\footnote{Ignoring or distorting them also reveals that there are ground conditions which are unforgotten against which such positions can be taken.} 

Some say that this is unavoidable
for any speech and action with implicature.
Pigeonholing this way of thinking to discourse-pragmatics or post-interpretation has not served us well either in language or in understanding action. 
In language for example, the reference intended (and let go) can affect grammaticality:

\ex\label{ex:plays}
She played the piano /(a) for an hour/(b) *in an hour/(c) in a year./
\xe

In (c), reference shifts from ordinary activity of play of (a) to master by practice. Consequently, the duration is not for performance, and completion
is for mastery. Like the ticket-fly example of planned action in the introduction, we can try to recover \emph{in an hour} of (b) as a period of learning for a genius, holding
to the shift of completion rather than one activity of play; but, this state of affairs also depends on the reference of \emph{play}, that is, event reference of a previously encountered element in the sequence. It is not simply implicature but categorial assumption about a reference (categorial because it affects grammaticality).

Using\xref{ex:plays} as a running example, the idea of complex category attempts to analytically address how we can avoid two extremes, syntacto-centric maximal generalism (e.g. one mental entry for all usages of \emph{play}) and cognito-centric polysemy fallacy,\footnote{\label{fn:fallacy}I borrow this term from \cite{sand:98}.}
(e.g. distinct mental entries for all distinct usages of \emph{play}). The common need for avoiding the idealistic extremes arises from the empirical notion of coping.

Language must go offline to be able to continue expressing ourselves, and planning cannot, otherwise we fail to empirically track the world.
Moreover, plan structures may reveal abstraction over predicates in the course of an action sequence, which in language would be the equivalent of \emph{movement} as transformational grammar employs to this day in addition to ``merge'' \citep{Chom:95}, ever since its inception by \cite{Chom:57}. 
It has been shown by \cite{stee:20} to be neither necessary nor sufficient  at level-2 for language. If we don't have to abstract over predicates
in language, what we have to abstract over is an important question in grounding thought and environment control.

This vantage point has implications for evolution of language from daily planning, that is, about Marr's level-3, which has been argued for by \cite{Stee:02,stee:17-ml}.
There is room for humble beginnings here for the most concrete level-3: 

\begin{quote}
The open-ended qualities of language go beyond signaling. The impetus for language has to do with wanting to ``tell'' someone else what is on our minds and learn what is on theirs. The \emph{desire} [my emphasis] to psychologically {connect}  with others had to evolve \emph{before} language. Only subsequently do the two sets of attributes coevolve.\\
\hspace*{25em}\citealt{hrdy:11}:38
\end{quote}
The colloquy of self and others moving about wanting to connect presupposes  co-location and interaction of self with others moving about, keeping track of each other's moves and intentions.

It has been very difficult to convince Chomsky and his colleagues that this is a simpler alternative to begin to understand evolutionary basis of language \citep{berw:16,haus:02,boec:07,boec:13,bolh:14}, rather than relying on a single biological event in the past, despite strong arguments in the first position's  favor \citep{toma:99,toma:09,Stee:02,stee:17-ml,hurf:12}. Taking that  route seems to  have been considered by Chomskyan practitioners to give diminishing scientific returns in discovering the fine structure and gaps in linguistic data. \cite{lieb:16}  attributes that
to staunch (and purportedly crypto) anti-Darwinism of Chomsky. I believe this is an unfair assessment.
I will go to some detail about this aspect.
The underlying reason for the Chomskyans' reluctance I think is the take on semantics.

\section{Hierarchical structures and locality of combination}\label{sec:baxter}

A mathematical result came from \cite{baxt:64}, now called \emph{Baxter  permutations}, from studying the invariants of hierarchical interpretability by  combination to a single result.
For a set of $n$ elements identified as $1, 2\cdots n$, he proved that
sequential ``patterns'' in the terminology of \cite{bose:98} which contain the subpattern $\cdot 3\cdot 1\cdot4\cdot2\cdot$ (i.e. sequences of elements
in this order, with possibly others in between shown by dots), 
and its mirror image $\cdot 2\cdot 4\cdot 1\cdot 3\cdot$, are the only ones which cannot be given
an overall local hierarchical interpretation. 

To give an impression of tree interpretation of a sequence and to explain why these two sequences stand out,  Baxter permutations are depicted by \cite{bose:98} as \emph{separation trees} where all leaves of   right binary trees are either less (-) or greater (+) than all leaves of left binary trees.\footnote{These trees are binary out of convenience; the values in them can be checked the same way in different trees. We shall see that binary combination is not a conceptual necessity, real or virtual; it is a mathematical result from
\cite{schonfinkel24}, with empirical significance for language and planning.} Examples   for $3124$ and  $2431$, which lead to possible hierarchies, are shown in the first column below. We only show $2431$ with many intervening elements for reasons of space; its pattern is in bold.
\ex\label{ex:sts}\footnotesize
\Tree [.+ [.- 3 [.+ 1 2 ] ] 4 ] \hfill~~~~~~~~~~~~~~~~~\Tree [.? [.+ [.- 3 1 ] 4 ] 2 ] \hfill \Tree [.? [.- 3 1 ] [.- 4 2 ] ] \hfill \Tree [.? 3 [.+ 1 [.- 4 2 ] ] ]

\Tree 
[.- [.+ [.+ {\bf 2} [.- [.+ [.- 5 {\bf 4} ] 6 ] {\bf 3} ]]  7 ] {\bf 1} ]
 \hfill \Tree [.? [.- [.+ 2 4 ] 1 ] 3 ] \hfill \Tree [.? [.+ 2 4 ] [.+ 1 3 ] ] \hfill \Tree [.? 2 [.- 4 [.+ 1 3 ] ] ]
\xe
We cannot separate trees involving $3142$ or $2413$. This is shown in second, third and fourth columns. Their overall results are marked `?' because of that.

\cite{stee:20} shows that in linguistic expressions of four or more elements these two are the only patterns
not attested in languages of the world. Only if attested they would force
the idea of movement (hence transformations), of the kind we see in Chomskyan grammar ever since its inception in 1957. \cite{stan:20} prove that Combinatory Categorial Grammar (henceforth \emph{categorial grammar} or CCG, depending on context) generates all and only the attested patterns, using application, composition (combinator \combb) and substitution (combinator \combs), as syntactically constrained combinators. 

The idea of combinator, mathematically a lambda term with no free variables, reflects our current mathematical understanding of how all combination can be made local and compositional. We shall see that they can help us reveal the fine structure in language and plans.
For example, the way we syntacticize the combinators is crucial in explaining the linguistic dimension of\xref{ex:sts}; see
\cite{candg,Bozsahin:12} for that.

These results do not suggest a return to dendrophilia \citep{Fitch:14}---our presupposed propensity to compute interpretations over a hierarchical structure---in analysis of an interpretation. They indicate that when we conceive semantics as
keeping track of and coping with change at every step (of the trees just shown, for example), then there must be a constraining mechanism that works transparently. It would
not generate the nonseparating trees in the first place, rather than generate and test them,
because it is a correspondence problem with an identifiable meaning at every step, with syntax (form and its structure) acting as  packaging of semantics (its content and structure).

Tree manipulation in this sense is not language-specific, and consequent use of natural recursion (which is recursion by value, not recursion by name) to build such trees is not human-specific \citep{bozs:16a}. Therefore mobile semantics has to say something more than interpreting the trees, because these  trees refer  to events which change.

A closer look at language and planning from this angle will reveal that assuming the same (but not identical ) representation is not only possible, it can go some ways in explaining the main difference between language and planning, that of offlineness and onlineness.\footnote{One other
obvious difference is the following: words are uttered, communicated and perceived, whereas plans are multimodal; they are acted, speech-acted, shared and received. In both cases,
the meaning of the behavior can only be guessed by the interlocutor in communication. It suffices for now to note that the constituents of
both kinds of behavior can arise from some form-meaning correspondence. I will be more explicit as we proceed.} 

The connection has already been made by \cite{Stee:02}, where it is argued that  combinator 
\combb\,is at the heart of composing compatible sequence of actions in an event calculus,
that is, for plan building; and, combinator \combt\,epitomizes affordances of \cite{Gibson:77}.\footnote{By definition, \combb=$\lambda\!f\lambda g\lambda x.f(gx)$, and \combt=$\lambda x\lambda f.fx$. One finding in linguistics is that all combining projections are ``curried,'' 
that is, they always take one function and one argument. In fact, \cite{schonfinkel24} had shown 16 years before Turing that all universal computation is ``Curried'', or, to undo some historical injustice, ``Sch\"onfinkelized''.
} 
\combb\,is Steedman's mechanism to carry over affordances
to exclusively backward chaining inference from the goal state to current state for more complex plans.

The roles of combinators in planning need not be pinned to specific constructions and composing action sequences. They seem to operate freely, with the exception of \combt. The findings which say that we need restrictions on \combt\,in action, just like in language \citep{stee:20}, suggest that a unified theory for plans and language is perhaps possible, and the common substrate of language and planning is more than we thought. 

There is one more result that we report which might be considered more technical: Once we \emph{lexicalize reasoning} with all its combinators involved for language and planning,
which is a  term I borrowed from \cite{geib:09,geib:15}, who use \combb\,only for sequence composition, following Steedman, we do not have to fix reasoning strategies in advance such 
as forward/backward reasoning. It corroborates further that similarities between language and planning are not coincidental.

\section{Syntax, semantics and local correspondences}

A major point of departure in current work from Chomskyan conception of semantics is that it is not taken as an appendix to syntax, or something that can be read off the syntactic trees.
It is based on the idea that the best we can hope for for understanding how simple constitution can give rise to complex behavior is to 
stick to the idea that syntax-semantics correspondence is the key point to understand, and doing syntax and semantics in lock-step in understanding a
linguistic expression
is a conservative way to start. Keeping the correspondences local, that is, compositional, and, interpertable by their own means, at the same time not disregarding potential nonlocal phenomena but not starting with them---in effect trying to explain nonlocal dependencies rather than capture them by auxiliary means, has payoffs in scaling up knowledge of language, and for understanding its connections to action.

Doing semantics with syntactic terms, or syntax with semantic terms, have not led to simple or transparent constrained mechanisms in linguistics, as 
the generative enterprise and cognitive linguistics have respectively shown.

Neither syntax nor semantics seems to suffice by themselves, but their correspondence can, in understanding the limits of linguistic structure. By making this move we would not be violating the autonomy of syntax, which is an important empirical claim that stood the test of time. As examples such as \emph{Colorless green ideas sleep furiously} show,
grammaticality does not mean having to make sense. However, the expression is meaningful, whereas nothing ungrammatical is meaningful. Therefore we are tasked with finding what is meaningful in senseless but grammatical expressions, with a solution
that can coextend to understanding the meaning of grammatical expressions with sense.
This is the 
stronger empirical basis of theories of correspondence.

The notion of transparent correspondence which gives us the leverage and courage to develop more and more complex models
 is our best bet at attempting to study acquisition of 
language \citep{Aben:15a}, also, organized behavior \citep{Mill:60}, and should not be abandoned in the first failure.

In a nutshell, categorial grammar proposes that
syntax is surface packaging of semantics. Although its ultimate
goal of semantics has been to understand change,\footnote{This has not been fully
incorporated into CCG yet, although the direction has been clear since \cite{Ades:82,Moen:88a,Stee:02,LewisM:13a,stee:19ohb}.} categorial grammar starts
with the task  of transparently projecting onto surface structures the syntax-semantics correspondences of elements of speech-sign, in a single computation, without any assumption of isomorphy in phonology syntax or semantics. Their structure is considered homomorphic to the unfolding
structure caused by analytics derived from CCG's principles implementing the language processor. As such, it is a type-2 theory. 
To integrate semantics as understanding how we keep track of change and cope with it in linguistic expression and daily planning, it takes truth-conditional semantics to be instrumental (or proxied) as long as it serves this purpose.
What we can say about content is not taken to be the central question, but the question of how we study it is. 

Looking at semantics this way cannot fall back
on hypotheses such as aforementioned \emph{dendrophilia} \citep{Fitch:14}---our presupposed propensity to compute interpretations over a tree i.e. hierarchical structure, which I do not question. Understanding and coping with change implies that semantics doesn't grow on syntactic trees otherwise syntax would suffice to do semantics in linguistics. It is the connection between the two and the limited degrees of freedom in correspondences which might give the dynamics of form-content idea some exploratory impetus.
Such dynamics does not appear to be domain-specific about meaningful sequences:
\begin{quote}
[T]he problems raised by the organization of language seem to me to be characteristic of almost all other cerebral activity. There is a series of hierarchies of organization; [..] not only speech, but all skilled acts seem to involve the same problems of serial ordering.\\
\hspace*{23em}\citealt{Lash:51}:121
\end{quote}
We need a vocabulary for the dynamics of action-by-doing and action-by-speech to support the idea of representing change itself. 

\section{Meaningful sequences}

We define \emph{meaningful sequences} as
succession of interpretable elements,  \emph{sequents}, whose types (sets of values) do not conflict in combining them in some way to a result---which is what makes them meaningful and interpretable.
For example, if we are given the sequence \emph{John ticket fly} as planned action, and asked to make altogether sense of it, we might
infer that John intends to fly and getting a ticket to fly is one way of enabling that, so he buys a ticket then flies. If we are told or we find out that the ticket is for a concert, we might infer that the concert might be located somewhere he needs to reach by flying. We change our understanding as we know more about the states of affairs and what these meaningful objects stand for, and whether the same sequence can be analyzed differently. This appears to be less haphazard than it sounds.

In the definition of meaningful sequences above we need to be explicit about `elements, types, combining' and `match'. In fact a more pressing need arising from the definition 
is understanding the commitment to sequences. 
Here we rely on the observation
mentioned earlier (\S\ref{sec:thought}) that all high level cognitive processes are sequential at the level they afford domain-internal inferences, although the physical substrate that supports them may be massively parallel. In this sense, and judging from \cite{Lash:51} and \cite{deac:97},
we seem to be \emph{homo sequens}.

We define \emph{mobile meanings}
to be channels of abstract communication for logical forms (LFs) in lieu of semantic representation, where the channels' spatiotemporality is determined by the participants of
the affair needing or causing the exchange and unfolding of the events in the affair. Such representations will be called Mobile Logical Forms (MLFs), to contrast with LFs as currently conceived in linguistics; see
\cite{May:85} for a review.

We define these terms as we proceed. Basically, a logical form is assumed (not unanimously)
in cognition
to be the place where thought is disambiguated \citep{Fodor:08}. An expression or overt behavior may be ambiguous, intentionally or unintentionally, also a belief or outcome of an action sequence, but there is believed to be a source from which these ambiguities can arise as a process. This justified or unjustified yet disambiguated defeasible representation
is assumed to be the basis of acquisition and learning by triangulation of inner and external world. As Fodor discussed at great length, an LF does not have to make
both thought and language compositional, but it must be that at least one of them is compositional to support the development of the other without infinite regress.\footnote{\cite{hume:echu} was aware of this problem long before
the idea of computer proposed an explicit mechanical solution for it, by
insisting on three atomic principles of connection in the mind.}

The constitutive elements 
are inherently sequential because they are shaped
by the principle of adjacency in CCG \citep{Ades:82}. Only two adjacent (i.e. overt) objects can be combined 
if they are meaning-bearing and combinable. 
If a sequence is not derivable by the projecting combinators of CCG,
then it is not meaningful. Any sequence derivable by them is meaningful, without an extraneous or nonlocal mechanism to make them meaningful.
These are the narrow claims of CCG.
Here `meaningful' means combinable by current means to be locally
interpretable, which we will clarify as we proceed.

The limits on the degrees of freedom in CCG categories are based on the idea of \emph{combinators}, which are, to repeat, lambda terms with no free variables. \cite{schonfinkel24} had thought of them as tools of universal
mathematical representation, which we now also know to be possible building blocks of universal
computation (see \cite{wolf:21} for a long list of century of contributions.)
We write them in bold, following the custom since \cite{curryfeys58}.
There are two of them in CCG. By definition, \combb=$\lambda\!f\!\lambda g\lambda x.f(gx)$, and \combs=$\lambda\!f\!\lambda g\lambda x.fx(gx)$.\footnote{Universal computation is captured by two combinators as well: \combs, defined above, and
\combk=$\lambda x\lambda y.x$. To informally make the connection to Turing machine, \combs\,expands and \combk\,shrinks the input, and \combs\,also composes.
As discussed in \cite{Bozsahin:12}, carving a linguistic theory out of combinators is an empirical effort because
not every possible computation and resource use is attested in natural language.}
Therefore we have $\combb abc$=$a(bc)$, and $\combs abc$=$ac(bc)$, given $a,b$ and $c$ in this \emph{applicative} (abstract, functional) order.
The organizational power of sequencing for semantics
comes from its inherent property of manipulating bound variables
differently in different \emph{empirical} order of combinations. For example $a,b,c$ above may be to the left or right in a given {empirical} order as long as they are given in the abstract order of first $a$, then $b$, then $c$, to \combb\,and \combs. This way sequents can rise above the meaning of individual elements themselves because the combiner can assign different meanings to abstract notions of `first' and `next', as well as to empirical notions of `left' and `right'.

CCG results indicate that sequencing itself, as in a sequence of words, 
if understood in combinatory categorial terms rather than as some linear metric, has a meaning that rises above the meaningful elements of the sequence themselves, to give rise to
more interpretable constituents with bounded and unbounded dependencies in the sequence \citep{steedman00}. 

Regarding the nature of dependencies in a linguistic sequence, clause structure seems to depend on thematic structure.
We have yet to find a syntax for some language that is oblivious to thematic structure
(agents, patients, experiencers, beneficiaries, etc.).

In linguistic discourse, i.e. connected clauses,
we do not see dependence on thematic structure within the connected clauses. It is mainly threading of clauses \citep{forbes06s}. We have not found discourses
that are not oblivious to thematic structure, although it is clear that rhetorical structure \citep{Mann:87} and conversational information structure \citep{Hall:70} are crucial in a discourse.

The differences between clausal syntax and cohesive communication bring
the study of relation between clauses and discourses on a par with relation between scripts and plans in organized action, which started quite early, for example \cite{Mill:60,scha:77}.
In planning, we do not see 
dependence of plans on participants until we think of them as arising from scripts implementing the plan, such as going to a restaurant, which is where thematic structure comes into play, as roles in a script. A plan requires a planner and a state of the world, to map onto another state of the world. A script is more specific than that because of assumed roles in its actions (e.g. clients, servers, waiters, queues, other guests, 
tables, 
etc.).
Unlike language, the semantics of planning and action must from the start cope with and determine the changing configurations in the world.

The new standard on planning, \cite{ghal:16},
continues to build on the earlier one by the same authors, \cite{ghal:04}, to promote the idea of  deliberate action as the new cornerstone of
planning research.\footnote{I will not use the terms `AI' and `planning' synonymously, not even coextensively, since
planning addresses a problem---organized behavior---whereas AI is a plethora of perspectives. The perspectives
were clear in the early AI, for example human extended-practice v. strong AI, but AI is now a nebulae of techniques 
whose intentions are no longer clear or clarifiable by its practitioners.} The contrast which is intended by the shift from planning-and-acting to planning-and-deliberately-acting is between  reactive and instrumental planning on one side, and deliberate planning on the other. The agent's view of the world takes the center stage
in understanding how they organize their actions to change the world. Plans lead to actions, and actions update and monitor the plans via participants, to give rise to \emph{what} and \emph{how} working together. Elaborate and diverse mechanisms have been proposed. This is a step toward a type-1 theory.

This line of thinking coincides with a similar development in linguistics, to the claim that languages differ only in their lexicons, and projection of lexicons
onto serial surface expressions is universal (that is, construction-free) \citep{steedman00,stee:20}. However, the connection is not evident.
These lexicons are correspondences of syntactic categories and meanings associated with words and similarly phonologically realized entities (such as signs), which embody how the agent constructs the linguistic world, and what is involved in the construction, by conceiving the nonlinguistic world and linguistic expressions, which is done with a procedural semantics. These meanings are \emph{mobile} in the simplest interpretation
of the term: the interpreter of the expression can change his mental linguistic world, both syntax and semantics, depending on further action. That in turn can change the world he lives in by acting on them. The connections to planning have already been noted by \cite{Stee:02,Stee:14e}.

This way of theorizing coming from linguistics adds another dimension to planning research, which is being refocused toward deliberation: Sequencing itself has a meaning, apart from the fact that elements in the sequence themselves have meanings. This perspective exports to planning the idea that although deliberative plans can deliver their meanings, by showing how they change the world, sequencing adds more ways to connect them which are not evident from  deliberate planners and deliberate plans themselves in isolation.

It will be argued that  procedural semantics must move up to mobile semantics to make the connection explicit.
In terms of evolution of language and planning, it probably progressed in the opposite direction, in which mobile meanings and online thinking in planning led to offline thinking, then to its expression, for more control. The quote from \cite{hrdy:11} at the end of \S\ref{sec:thought}  is intended with this connection in mind. It is also the idea behind the conjecture that language is expression of thought.\footnote{The current
linguistic landscape is more complex than this sloganish statement, which I offer  by way of summary. There are narrow constraints
in the way thought can be expressed in language, because there are gaps in linguistic data. We will cover some of them.

\citealt{Fodo:75}:84 cites \cite{brya:74} about the priority of thought to language in light of work coming from language acquisition research, which is worked the other way around in various forms of linguistic relativism. The circularity of the argument ``[that children begin to take in and use relations to help them solve problems because they learn the appropriate comparative terms like `larger'] generalizes, with a vengeance, to any proposal that the learning of a word is essential to mediate the learning of the concept that the word expresses.''}

{Mobile meanings}
are changing configurations in the world. 
I adapt the term from Robin Milner's lifetime work  in computer science on mobility and its semantics, summarized in his Turing lecture 
on $\pi$-calculus \citep{miln:93}. He showed in \cite{miln:92} that what $\lambda$-calculus does to sequential meanings, namely capturing mathematics of communicating an argument to a function, is a special case  of what $\pi$-calculus does to mobile or communicating concurrent processes.
 
The relevance of these aspects of mobility to planning is clear but not obvious, and connections to language need a detailed argument. We shall make the two-way connection explicit to first build a joint theory at level-2. It has implications about what level-1 would need from such a theory.

In particular, it will be argued that
if plans are states of a process, they must refer to states of the world rather than participants of the world. (Something else, namely a script, refers to participants in the course of a plan.) 
In language, we observe no natural language in which clauses can be embeddingly built without thematic structure, that is, without reference to participants, and with reference only
to states of the world. This has implications for level-2 support for level-1 in a type-1 theory.

If thematic structure can only be read off the logical forms of meaning-bearing elements (of words and actions), because it is not in the states of the world, and plan states cannot be read off the linguistic logical form because
the form only delivers participants,
then language can ignore the conditions of the world but planning cannot. If plans must refer to conditions of the world, 
whereas language cannot retrieve participants from that structure,
then what is left for language is only sequentially structured meanings, roughly who-does-what-to-whom-where-and-when, but not how. These concepts
of a level-2 argument have to be made explicit, which we do in next three sections.

One further implication of this way of thinking is that language is serial not only because of phonology, but also because of syntax. And, planning need not be serial, because of its semantics, not syntax. However, something serializes it.
This concurs
with the principle adopted here, that order gives rise to structure, rather than the other way around. The categorial conception of order cannot be reduced to linear metrics like distance or weight, as we shall see in \S\ref{sec:language}.

A fundamental level-2 question that seems to arise from joint consideration of language and planned-collaborative action appears to be the following: How do we prepare ourselves to cope with change, assuming offlineness of language of online-acting agents, and onlineness of daily activity of expressively offlined agents? This question connects to the fundamental level-1 question: What is all that preparation for? Why do we express thought?

The rest of the article proceeds as follows. Section~\ref{sec:plans} introduces some developed vocabulary in trying to cope with change, and builds sequential categorial inference on it for joint consideration of syntax and semantics of plans.
Section~\ref{sec:mobile} introduces mobile semantics, to turn terms of change into fluents. Section~\ref{sec:language} reiterates current developments in linguistics for joint consideration of syntax and semantics. Section~\ref{sec:synthesis} elaborates on the implications of the terms of the synthesis. We conclude with a discussion (Section \ref{sec:conclusion}).

\section{Plans}\label{sec:plans}
Plans are binary functions, taking a planner and states of the world to another state of the world, specifying \emph{what} does that. \emph{Scripts} however, or in more recent technical planning terminology, operators, heuristics and task networks, deal with \emph{how}, and necessarily involve thematic relations, engendering states of the world only as a consequence. The term `script' was borrowed by computer scientists from cognitive psychologists of 1960s \citep{scha:77}.

Consider one classical example of  planning, from \cite{sace:75a,sace:75}: painting a ceiling and painting a ladder. 
An agent picks up a ladder and some paint, and paints. One way to describe the changing states of this micro-world is the following sequence of terms and actions:
\ex\label{ex:sace}
Ed\\
ceiling\\
paint\\
Ed\\
ladder\\
climb-up\\
paint\\
climb-down\\
paint
\xe
Although it is ambiguous, one cannot help but interpret the sequence as: Ed wants to paint the ceiling, picks up some paint, gets himself a ladder,
climbs it, and paints the ceiling. Then he comes down and paints again, this time perhaps the ladder. 

Ed could have painted the ceiling first somehow, climbed the ladder, and picked up a paint, then come down and painted the ladder. This is not the likely sequence of affairs, but it is possible. The scenario above would be missing some parts in this case.
What is less likely, \emph{given this sequence}, is painting the ladder first, then the ceiling, because of the changing configurations (states of the world) and what they mean to a human being. The change of states of affairs is easier to follow in narrower circumstances.

In the simple blocks-world, which is another planning classic \citep{Wino:72}, we can think of a state of affairs where block B2 shown on the right in\xref{ex:blocks} is on top of block B1, which is on table T1. The  sequence on the left cannot be interpreted as picking up B1, \emph{given the world on the right}:\footnote{\cite{Wino:72} used the blocks-world
to narrowly control instructions in natural language to change the world, as demonstration of understanding the expressions. We are running the blocks-world without linguistic instructions first.}
\ex\label{ex:blocks}
\begin{minipage}[t]{7ex}
John\\
B1\\
pick-up
\end{minipage}\hfill\begin{minipage}[t]{7ex}\raisebox{-3em}{\includegraphics[scale=.6]{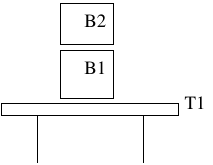}}\end{minipage}\hfill\ 
\xe
It might mean John wanting to pick up B1, which requires a scripted plan to carry it out.
Now consider the following sequence:
\ex\label{ex:2}
ceiling\\
paint\\
Ed
\xe
It might mean that the ceiling needs painting, provided we find an agent further down the scenario, say Ed. (It might also mean someone first reaches the ceiling, then paints Ed, but this seems less likely---yet meaningful.) We have a propensity to want to  make sense of things no matter how unlikely they are, if they are combinable somehow. This was the point of Chomsky's \emph{colorless green ideas sleep furiously} example for language.

These interpretations follow from an understanding that if A is established earlier it might enable state B, as in~(\ref{ex:sace}--\ref{ex:blocks}), and reach something
meaningful when combined,
provided that its conditions meet the current situation related to B, which is not the case for the first interpretation of\xref{ex:blocks}.

It might also be the case that A can lead to expect a state B, as in\xref{ex:2}. Connections to forward and backward chaining in planning research
are worth pointing out (respectively: going from current state to goal state, or from goal state to current state). Given the inherent temporal asymmetry of sequencing, and without reference to physical time, we can surmise the following: If A enabled B, then we can perhaps reach B from A. If A leads one to expect B, then we reason backwards from B to see if A is realizable from thinking B as a possible outcome.\footnote{As an exposure to mechanics of planning, \cite{russ:10} provide
a basic coverage of planning strategies. \cite{ghal:04,ghal:16} are definitive collections on planning.
For a  survey of planning, see \cite{geff:13}.} The difference has everything to do with being able to think about the past and future and doing something about it by combining.
We build a representational base bearing these properties in mind.

\subsection{Sequences and plan categories}\label{sec:sac}
No matter in which direction the states of affairs unfold, we have to face the Frame Problem, which is deciding what changes and what remains the same during actions \citep{McCa:69}, which is one of the most important contribution of computer science to philosophy.\footnote{Another one, in my opinion,
is being explicit about how to avoid the Cartesian Theater of \cite{denn:91}, which is concerned with the infinite regress of interpreting the world without recursion-stopping and translation-stopping primitives, one solution coming from theory of programming languages in the concept of `virtual machine'.}

We begin to make explicit the sequence asymmetry  using the STRIPS mechanism of \cite{Fike:71}. STRIPS is nowadays considered hopelessly simplistic and anomalous to face real planning tasks, but
it is a good start.
It captures the Frame Problem so well that it served both as the grandfather and subset of most---if not all---modern planning systems. 
This is because its procedural semantics fits well with the nature of the problem: every sequence of actions introduces a side effect to the world, which might be a realization of a state, or give rise to an expectation that will lead to a state.\footnote{We shall, however, not use the stack-based plan decomposition mechanism of STRIPS and similarly inspired offshoots,
which is known to be susceptible to incomplete and even undecidable computations when a simple solution
is available. The simplest example of this kind is the \cite{suss:75} anomaly.}

To make the discussion more concrete, we give the following abstract spatiotemporal (i.e. sequential) semantics to symbolizing  sequences. 
\pex\label{ex:seqcat}
\a $\alpha$  \plancat \cgf{X\bs Y}\\
It means that category \cgf{Y} of an earlier sequent (notice `\bs') may enable sequent $\alpha$, the bearer
of \cgf{X\bs Y}, to reach \cgf{X}.

\a $\alpha$ \plancat \cgf{X\fs Y}\\
It means that category \cgf{Y} of a later sequent (notice `\fs') may give rise to \cgf{X}
for sequent $\alpha$, the bearer of \cgf{X\fs Y}.
\xe
In $\alpha \plancat \beta$, $\alpha$ is the bearer of category $\beta$; $\alpha$ can be an agent or subject holding a category for certain action; it can also be a property ascribed by agents and subjects for the affordance of some action abstracted in $\beta$.
In\xreff{ex:seqcat}{a}, \cgf{Y} is expected to be before \cgf{X\bs Y} because of `\bs'. In\xreff{ex:seqcat}{b}, \cgf{Y} is expected to be after \cgf{X\fs Y} because of `\fs'.

The sequents can be anything meaningful in a sequence, such as\xref{ex:sace}. 
The modal \emph{may}  in the expressions of\xref{ex:seqcat} is  because of the fact that we do not know what kind of plans \cgf{X} and \cgf{Y}  are part of. Once we fix the plan, that is, \cgf{X} and \cgf{Y}, or more accurately, the script, the modals can be dropped, as we shall see.

The sequential asymmetry of this semantics can be shown as follows, where e.g. the sequence \cgf{A~B} means \cgf{A} is before \cgf{B}, and `$\not\Rightarrow$' means `not possible'. 
\ex\label{ex:pl-fa}
\begin{tabular}[t]{llll@{\hspace{18em}}r}
$\alpha$ \plancat  \cgf{Y}  &$\beta$ \plancat  \cgf{X\bs Y}& $\Rightarrow$ &$\alpha\beta$ \plancat  \cgf{X} & (apply)\\
$\alpha$ \plancat  \cgf{X\fs Y} &$\beta$ \plancat  \cgf{Y} &$\Rightarrow$  &$\alpha\beta$ \plancat  \cgf{X}& (apply)\\
$\alpha$ \plancat  \cgf{X\bs Y}&$\beta$ \plancat  \cgf{Y}  & $\not\Rightarrow$ &$\alpha\beta$ \plancat  \cgf{X}\\
$\alpha$ \plancat  \cgf{Y} &$\beta$ \plancat  \cgf{X\fs Y} & $\not\Rightarrow$  &$\alpha\beta$ \plancat  \cgf{X}
\end{tabular}
\xe
The third line says that enabling requires prior occurrence of the enabler. The fourth line says that creating an expectation means it has not been realized 
yet.\footnote{\cite{geib:09} gives the backslash the semantics of `earlier than' and the forward slash `later than'. We shall see
that mobile semantics can deliver the  spatiotemporality of this semantics without these assumptions, leaving the intensional interpretations to the plan itself. In other words, these categories are intensional objects
associated with a sequent. Mobile semantics will supplant
Steedman's (\citeyear{Stee:02}) use of dynamic event calculus of time points for sequencing but shares with it the adoption of STRIPS semantics in addressing the Frame Problem.} The last two lines show that  the space of representations is limited, and asymmetric.

There is no ``action at a distance'' in these combinations, and all  combination is local. This is what it means to be in a sequence which is meaningful by itself.
For example the following combinations are not possible (dots denote some sequentially
intervening nonempty action resources). The intervening
material itself has to connect for combination in\xref{ex:pl-fa} to happen.
\ex\label{ex:pl-fa2}
\begin{tabular}[t]{lllll}
$\alpha$ \plancat  \cgf{Y} &$\cdots$ &$\beta$ \plancat  \cgf{X\bs Y}& $\not\Rightarrow$ &
$\alpha\beta$ \plancat  \cgf{X}\\
$\alpha$ \plancat  \cgf{X\fs Y} & $\dots$ &$\beta$ \plancat  \cgf{Y} &$\not\Rightarrow$  &
$\alpha\beta$ \plancat  \cgf{X}\\
\end{tabular}
\xe
This property is essential to make the whole sequence locally connected, equivalently, meaningful by itself, if it can be combined.

Enabling in\xref{ex:seqcat} does not mean applicability. It makes no reference to current states of the world. The same is true of leading;  future states caused by \cgf{Y} are not referred to. 
STRIPS semantics and representations can help  clarify such properties.

\subsection{Applicative structure of plans}
To implement the semantics of directionality in a sequence in\xref{ex:seqcat}, which is represented by slashes, we complement the system with two substantive categories on the combinatory side to match placeholders \cgf{X} and \cgf{Y} from abstractions about the world:
\ex
\cgf{S}, corresponding to \emph{states},\\
\cgf{T}, corresponding to \emph{terms}. 
\xe
These are extensive enough
to be able to abstractly conceive various correspondences of actions and events.  For example, 
some terms are agents, such as  the planner, but not all, such as a falling tree.  Some states are terms, such as holding a ticket for a flight, but not all, such as boarding a plane, which is a process. 

\cgf{X} and \cgf{Y} above generalize over substantive categories \cgf{S} and \cgf{T} and their combinations. Nothing in the discussion hinges on having a universal
repertoire of substantive categories; they can be extended from the minimal set of \cgf{S} and \cgf{T}.

Consider the act of picking something up in the sequential configuration of\xref{ex:blocks}, this time in more detail using the vocabulary just introduced:
\ex\label{ex:pickup}
\begin{tabular}[t]{lll}
pick-up & \plancat  & \cgf{(S\bs T)\bs T} $:\lambda x\lambda y\lambda z.\so{pickup}xy$\\
&& \texttt{pre:} $inhand(y,nil), clear(x), block(x), on(x,z)$\\
&& \texttt{add:} $inhand(y,x), clear(z)$\\
&& \texttt{del:} $inhand(y,nil), on(x,z)$
\end{tabular}
\xe
\cgf{(S\bs T)\bs T} is the category which says that two terms (\cgf{T}) give a state (\cgf{S}), by enabling one by one. In other words, in sequence symbolism,
it is a function from outer `\cat{\bs T}' to \cat{(S\bs T)}, which is itself
a function from  `\cat{\bs T}' to \cat{S}. In other words, empirical order is a property of the input rather than the output.
The plan's argument/operator structure
is written as a lambda term after the colon, whose effect is to realize the conditions in the \texttt{add} list, and cancel the conditions in the \texttt{del} (delete) list, provided its preconditions in the \texttt{pre} list are satisfied in the states of affairs it applies. This is from STRIPS. I follow STRIPS semantics for the time being and \cite{Pietroski:16a} in also assuming that all conditions are conjunctions.

The first term to combine with pick-up action is the outermost \cgf{T}, corresponding to $x$ in the operator because of lambda correspondence. Once we make the correspondence, it can be inferred to be not the doer of picking but what is picked, because of its role in the predicates of\xref{ex:pickup}. We can infer this structurally, i.e. without labels,  if we assume that the predicate \so{pickup} has a structured representation where
more prominent arguments command less prominent ones, as below (the prime notation indicates semantic predicate):\footnote{It is important
that any notation disambiguates the correspondence of lambda bindings and action types. This notation achieves that, and its unusual choice of representing command relations will be defended on linguistic grounds later. }
\ex\label{ex:pickup-tree}
\Tree [[ \so{pickup} $x$ ] $y$ ]
\xe
Thus we do not need conditions such as $agent(y)$ or $patient(x)$ to infer relative prominence. All we need is an interpretation of structural asymmetry.

Specifications such as\xref{ex:pickup} are not just plans, which are functions from states to states,
but scripted plans, or scripts for short. They have thematic structure because of the roles determined by correspondence.\footnote{Although these labels  are borrowed from \cite{scha:77}, we do not make representational
distinctions between plans, scripts, goals and themes as they do. The narrower claim here is that all planned activity
can be organized with this mixture of category, argument structure and conditions defined uniquely per planning object's sense, allowing for example scripts
to have subplans as arguments, with their own script, and themes to differ depending on order and type of arguments, and so on.
One computational consequence of this representational simplification is that probabilistic inference involved in learning need not know anything about their source, which significantly simplifies the learning model. Connections to language learning is covered later.}  \cgf{T} (term) does not carry with it a thematic role unless its prominence is bound
to a corresponding lambda term. Likewise for \cgf{S} (state).  

The doer of the action in\xref{ex:pickup}, $y$, is also inferred structurally. It is the most prominent argument in\xref{ex:pickup} because
$\so{pickup}xy$ is equivalent to $((\so{pickup}x)y)$ by left-associativity of juxtaposition. Sequencewise it corresponds to
the inner term \cgf{T} in\xref{ex:pickup}, so that John is the picker-upper and B1 is the target in a sequence like\xref{ex:blocks}.

The term $z$ in\xref{ex:pickup} is not part of the sequence semantics per se  because it does not correspond to an argument term or state of the plan. It allows participants to refer to the state of the world so that the action of pick-up can be compositionally specified. Notice that what is picked up, $x$, is on top of something which is cleared, and this property may prove essential for future acts, which is threaded through the arguments of the sequence by $z$. In the context provided for the discussion of\xref{ex:blocks} it is the block B1.

With the explicit representation in\xref{ex:pickup} we can infer why  sequence\xref{ex:blocks} cannot mean John picks up block B1:
$clear(\so{b1})$ is false at the moment the plan is applied. (For example, it may have been falsified by the {\tt del} list
of a previous put-on action, e.g. putting B2 on top of B1.)\footnote{Here we assume, following STRIPS, that literals which are not listed in the world state are false.
Deleting $clear(\so{b1})$ thus achieves the desired effect using this closed-world assumption.}
Therefore the plan fails and the world stays the same for the planner.

If the sequence were John-B2-pickup under the same state of the world where B2 is on top of B1, the plan realization would proceed as follows where we write the timeline/sequencing from left to right and plan unfolding from top to bottom.
\ex\label{ex:jb2p}
\cgex{3}{John  & B2 & pick-up\\
\cgul & \cgul & \cgul\\
\cgf{T}$:\so{john}$ & \cgf{T}$:\so{b2}$ & \cgf{(S\bs T)\bs T}$:\lambda x\lambda y\lambda z.\so{pickup}xy$\\
\ml{1}{\texttt{pre:}-} &\ml{1}{\texttt{pre:}-} & \ml{1}{\texttt{pre:} $inhand(y,nil), clear(x), block(x), on(x,z)$}\\
\ml{1}{\texttt{add:}-} &\ml{1}{\texttt{add:}-} & \ml{1}{\texttt{add:} $inhand(y,x), clear(z)$}\\
\ml{1}{\texttt{del:}-} &\ml{1}{\texttt{del:}-} & \ml{1}{\texttt{del:} $inhand(y,nil), on(x,z)$}\\
& \cgline{2}{apply}\\
& \cgres{2}{S\bs T\lf{\lambda y\lambda z.\so{pickup}\so{b2}y}}\\
&\ml{2}{\texttt{pre:} $inhand(y,nil), clear(\so{b2}), block(\so{b2}), on(\so{b2},z)$}\\
&\ml{2}{\texttt{add:} $inhand(y,\so{b2}), clear(z)$}\\
&\ml{2}{\texttt{del:} $inhand(y,nil), on(\so{b2},z)$}\\
\cgline{3}{apply}\\
\cgres{3}{S\lf{\lambda z.\so{pickup}\so{b2}\so{john}}}\\
\ml{3}{\texttt{pre:} $inhand(\so{john},nil), clear(\so{b2}), block(\so{b2}), on(\so{b2},z)$}\\
\ml{3}{\texttt{add:} $inhand(\so{john},\so{b2}), clear(z)$}\\ 
\ml{3}{\texttt{del:} $inhand(\so{john},nil), on(\so{b2},z)$}\\
}
\xe
The overall plan succeeds because its ground conditions reached at the bottom are met, hence the world is updated.
The partial plan, which is the first application in the middle line, partially succeeds as well because its grounded preconditions were not violated.\footnote{Ground instances have either atomic arguments like \so{b2} or atomizable arguments
such as $z$ in \cgf{S\fs T}$:\lambda x\lambda z.\cdots \mbox{\tt pre:} on(x,z)$, where $z$ is not an argument of the plan (unlike $x$), therefore
it can be existentially substituted for something, say \so{a}. We differ from similarly STRIPS-inspired probabilistic
planners such as \cite{Pasu:07} in making
use of potentially nested function symbols
in groundable instances, rather than flat relational structures. Functional STRIPS of \cite{geff:00} addresses such semantics
computationally.}
It can be paraphrased as `pick-up B2 plan'. Because its plan arguments were not all grounded or groundable, it has not been executed yet.
Thus plan failure can be captured as early as the sequencing would allow.

The logic of requiring all preconditions to be groundable before we change the world is that the situation might change if the plan details are not yet completed. For example, before John attempts to pick up B2, something may have been put on top of B2. If this happened
in between John and B2, then $clear(\so{b2})$ would be false (deleted) \emph{when} John is incorporated into the plan.

This is an instance of  forward chaining (going from state to goal) because the current world set up by B2 and John is fed into pick-up. It shows that, unlike STRIPS, which is strictly backward chaining from a goal state to current state, forward or backward chaining does not have to be a fixed strategy in planning once it is scripted.\footnote{We shall see later that plan structures such as\xref{ex:jb2p} are only apparently backward chaining because of the appearance that the plan goes back to its participants. The idea of type-raising the terms
to look for plans they can participate in goes back to Gibson's (\citeyear{Gibson:77}) affordances, and it is crucially
involved in Steedman's (\citeyear{Stee:02}) account of plans and grammars. The point to make here is that
type-raising John and B2 can begin to look for a goal state which satisfies their affordances, therefore this example is
not\emph{ necessarily} backward application.
We shall say much more about the connection later.}
 The order and categories can facilitate either one. Therefore it depends on the plans and the sequence.
 The upshot of that for the developing proposal is that we must allow for any kind of thought or reasoning to be expressed so that level-1
 can  be concerned with its environment control.
Here is an example which requires backward chaining:
\ex\label{ex:back}
pick-up\\
B2\\
John
\xe
This sequence seems to suggest that pick-up raises an expectation for something to be picked up, by somebody. The current state
which pick-up sees is not the relevant precondition, therefore this is more like backward chaining (from goal to state), to states of the world to be created by B2 and John from the expectation of pick-up.

We conjecture that this is always caused by \cgf{X\fs Y} kind of plan categories in\xref{ex:seqcat}, requiring information about future conditions to realize the goal, so that it always has to start with a known goal \cgf{X} because of the inherent asymmetry of sequencing, rather than being an order-free option for  planning. The argument in the article is that there is no need for a separate planning algorithm at level-2 if all combinations depend on the plans and their sequencing.
This is in fact a level-1 concern.

Therefore pick-up of\xref{ex:back} is categorially different than pick-up of\xref{ex:blocks}. 
These differences, and the explicit demonstration of backward chaining from the new category of pick-up, are shown below; cf.\xref{ex:jb2p}.
\ex\label{ex:pjb2}
\cgex{3}{pick-up & B2 & John\\
\cgul & \cgul & \cgul\\
 \cgf{(S\fs T)\fs T}$:\lambda x\lambda y\lambda z.\so{pickup}xy$
&  \cgf{T}$:\so{b2}$ & \cgf{T}$:\so{john}$ \\
\ml{1}{\texttt{pre:} $inhand(y,nil), clear(x), block(x), on(x,z)$}&
\ml{1}{\texttt{pre:}-} &\ml{1}{\texttt{pre:}-}\\
\ml{1}{\texttt{add:} $inhand(y,x), clear(z)$}& \ml{1}{\texttt{add:}-} &\ml{1}{\texttt{add:}-} \\
\ml{1}{\texttt{del:} $inhand(y,nil), on(x,z)$}&\ml{1}{\texttt{del:}-} &\ml{1}{\texttt{del:}-} \\
\cgline{2}{apply}\\
\cgres{2}{S\fs T\lf{\lambda y\lambda z.\so{pickup}\so{b2}y}}\\
\ml{2}{\mbox{\tt pre:} $inhand(y,nil), clear(\so{b2}), block(\so{b2}), on(\so{b2},z)$}\\
\ml{2}{\mbox{\tt add:} $inhand(y,\so{b2}), clear(z)$}\\
\ml{2}{\mbox{\tt del:} $inhand(y,nil), on(\so{b2},z)$}\\
\cgline{3}{apply}\\
\cgres{3}{S\lf{\lambda z.\so{pickup}\so{b2}\so{john}}}\\
\ml{3}{\mbox{\tt pre:} $inhand(\so{john},nil), clear(\so{b2}), block(\so{b2}), on(\so{b2},z)$}\\
\ml{3}{\mbox{\tt add:} $inhand(\so{john},\so{b2}), clear(z)$}\\
\ml{3}{\mbox{\tt del:} $inhand(\so{john},nil), on(\so{b2},z)$}\\
}\xe
We may question why this sequence is necessarily backward chaining from goal state. After all, we could reach a combined expectation from B2 and \emph{John}, say as something John does to B2. However, pick-up is \emph{earlier} in the sequence,\footnote{It does not follow that it \emph{happened} earlier \emph{because} it is earlier in the sequence. Notice that pick-up's category is not a proposition but a predicate, something in need of arguments.}
unlike\xref{ex:jb2p}, hence the
goal is already set up, needing its participants. It would be incongruous with the assumption of meaningful sequences to disregard the expectation of an early appearing plan, find some terms later that can have affordances about an earlier plan, and start worrying then about what the plan might be. As we shall see later, the situation in\xref{ex:pjb2} is different than having a plan early and its arguments even earlier.
Level-2 must make this distinction so that level-1 can only depend on what the problem space is, rather than how it is represented
or solutions are sought.

This way different ways of realizing pick-up can be distinguished in representation and inference, in goal-directedness and situation awareness.

Notice also that everything is a plan, including terms like John and B2, not just actions or states, because they all have
a category, an operator, precondition, add and delete lists. This property will help us conceive all sequences as scripted plans.

\subsection{Rising above applicative structure of plans}

So far we have seen the simple applicative structure of plans, where for example applying \cgf{(S\bs T)\bs T} to a \cgf{T} before it yields \cgf{S\bs T}, with concomitant substitution for the participants involved and conditions to be checked. However,
such simple state of affairs might suggest that planning behaviour is complex only because there can be plans with many arguments and subplans. It can also be complex because of argument-sharing and chaining across plans.
The added predictive behavior comes from semantics of sequencing.

For example, the following sequence can be interpreted in several ways. (From now on we write sequents from left to right to save space.)
\ex\label{ex:fly}
John fly buy  ticket
\xe
We can think of \emph{fly} as an act involving an instrumental use. That is,
John takes the flight in a plane by certain means rather than fly the plane himself.  

It is also possible that the flier and the ticket buyer are not the same person. To be able to guess correct sense of questions such as \emph{Who is John's travel agent?} or \emph{How did John fly?} we must make the sequence's structure explicit in this interpretation. Figure~\ref{fig:jfbt} is one way to do it.

\begin{figure}[h!]
\footnotesize
\cgex{4}{John & fly & buy & ticket\\
\cgul & \cgul & \cgul & \cgul\\
\cgf{T} & \cgf{(S\fs S)\bs T} & \cgf{(S\fs T)\fs T} & \cgf{T}\\
$:\so{john}$ & $:\lambda x\lambda s\lambda z.\so{takeflight}\,s\,z\,x$ & $:\lambda x\lambda y\lambda z.\so{buy}zxy$ &$:\so{ticket}$\\
\ml{1}{\tt pre:-}  & \ml{1}{{\tt pre:} $able(x,s),flight(z)$} & \ml{1}{{\tt pre:} $payable(x),funds(F),have(y,F)$} &\ml{1}{\tt pre:-}\\
\ml{1}{\tt add:-} &\ml{1}{{\tt add:} $at(x,dest(z)),at(z,dest(z))$}&\ml{1}{{\tt add:} $have(y,x),able(y,z)$}&\ml{1}{\tt add:-}\\
\ml{1}{\tt del:-} &\ml{1}{{\tt del:} $at(z,orig(z)), at(x,here)$} &\ml{1}{\tt del:~-} &\ml{1}{\tt del:-}\\
\cgline{2}{apply}\\ \cgres{2}{S\fs S}\\
\cgres{2}{\lf{\lambda s\lambda z.\so{takeflight}\,s\,z\,\so{john}}}\\
\ml{2}{{\tt pre:} $able(\so{john},s),flight(z)$}\\
\ml{2}{{\tt add:} $at(\so{john},dest(z)),at(z,dest(z))$}\\
\ml{2}{{\tt del:} $at(z,orig(z)), at(\so{john},here)$}\\
& &\cgline{2}{apply}\\ &&\cgres{2}{S\fs T}\\
&&\cgres{2}{\lf{\lambda y\lambda z.\so{buy}\,z\,\so{ticket}y}}\\
&&\ml{2}{{\tt pre:} $payable(\so{ticket}),funds(F),have(y,F)$}\\
&&\ml{2}{{\tt add:} $have(y,\so{ticket}),able(y,z)$}\\
& &\ml{2}{{\tt del:-}}\\
\cgline{4}{compose}\\ \cgres{4}{S\fs T}\\
\cgres{4}{\lf{\lambda x\lambda y\lambda z.\so{takeflight}(\so{buy}\,z\,\so{ticket}x)\,y\,\so{john}}}\\
\ml{4}{{\tt pre:} $able(\so{john},\so{buy}(z,\so{ticket},x)),flight(y), payable(\so{ticket}),funds(F),have(x,F)$}\\
\ml{4}{{\tt add:} $at(\so{john},dest(y)),at(y,dest(y)),have(x,\so{ticket}),able(x,z)$}\\
\ml{4}{{\tt del:} $at(y,orig(y)), at(\so{john},here)$}\\
}
\caption{A plan structure for John flying after someone will have bought a ticket.}
\label{fig:jfbt}
\end{figure}

Notice that \emph{fly} does both forward and backward chaining. John is the flier, and he needs a state where the ticket is bought by him or by someone else, therefore it is a two-argument plan, more accurately a two-argument script, involving one term and one state. \emph{Buy} can expect a buyer, and given these specifications, \emph{ticket} can be followed by John or someone else as the buyer. 

This is function composition, composing
\cgf{S\fs S} of John-fly with \cgf{S\fs T} of buy-ticket. These are considered valuations of the
placeholders \cat{X, Y, Z}:
\ex\label{rul:fc}
\cat{X\fs Y}\lf{f}~~~\cat{Y\fs Z}\lf{g}~~$\Rightarrow$~~\cat{X\fs Z}\lf{\lambda x.f(gx)}\hfill(\combb)
\xe
If the first category were \cgf{S\fs T} for John-fly, meaning
John flies something rather than flies in something after attaining a certain state, the composition couldn't have happened. It is in this sense that plans, equivalently states, are category-dependent. That is what makes them interpretable.

Now consider the case in which the sequence\xref{ex:fly} does not give rise to an expectation that someone else will buy the ticket,  and John is not expected to \emph{appear} again in the sequence but stay involved in the whole scene. Say, he will have bought the ticket himself by the end of the sequence, and this was all part of the plan since his first appearance.
John appears once in the sequence, and does two acts, buying and flying. 

This possibility is not an intrinsic property of the plans involved. It can be revealed by the semantics of sequencing.
It differs from plain composition of\,$f$\,and\,$g$, viz. $\lambda x.f(gx)$, although they are related. Its semantics is category-dependent too, as follows:
\ex\label{rul:sub}
\cgf{(X\fs Y)\bs Z}$:f$~~~\cgf{Y\bs Z}$:g$ $\Rightarrow$ \cgf{X\bs Z}$:\lambda x.fx(gx)$\hfill(\combs)
\xe
\cite{schonfinkel24} invented its mathematics and called  it \emph{fusion}, which \cite{curr:29} renamed as \emph{substitution}, which is the name currently in use.

Because John is not expected to reappear in the sequence in this interpretation, which means 
he is the buyer of ticket and the flier, the plan category for \emph{buy} ought to be different, say \cgf{(S\bs T)\fs T}, where the `\bs T' is John, appearing earlier.
(This is an implicit way to say that plans are resource-sensitive, because every appearance in a sequence counts.) These states of affairs unfold in the plan structure of Figure~\ref{fig:subsplan}. This sequence's semantics is the one that makes use of\xref{rul:sub}.

\begin{figure}[h!]
\begin{center}\footnotesize
\rotatebox{90}{
\cgex{4}{John & fly & buy & ticket\\
\cgul & \cgul & \cgul & \cgul\\
\cgf{T} & \cgf{(S\fs S)\bs T} & \cgf{(S\bs T)\fs T} & \cgf{T}\\
$:\so{john}$ & $:\lambda x\lambda s\lambda z.\so{takeflight}\,s\,z\,x$ & $:\lambda x\lambda y\lambda z.\so{buy}zxy$ &$:\so{ticket}$\\
\ml{1}{\tt pre:-}  & \ml{1}{{\tt pre:} $able(x,s),flight(z)$} & \ml{1}{{\tt pre:} $payable(x),funds(F),have(y,F)$} &\ml{1}{\tt pre:-}\\
\ml{1}{\tt add:-} &\ml{1}{{\tt add:} $at(x,dest(z)),at(z,dest(z))$}&\ml{1}{{\tt add:} $have(y,x),able(y,z)$}&\ml{1}{\tt add:-}\\
\ml{1}{\tt del:-} &\ml{1}{{\tt del:} $at(z,orig(z)), at(x,here)$} &\ml{1}{\tt del:-} &\ml{1}{\tt del:-}\\
& &\cgline{2}{apply}\\ &&\cgres{2}{S\bs T}\\
&&\cgres{2}{\lf{\lambda y\lambda z.\so{buy}\,z\,\so{ticket}y}}\\
&&\ml{2}{{\tt pre:} $payable(\so{ticket}),funds(F),have(y,F)$}\\
&&\ml{2}{{\tt add:} $have(y,\so{ticket}),able(y,z)$}\\
& &\ml{2}{{\tt del:-}}\\
&\cgline{3}{substitute}\\ &\cgres{3}{S\bs T}\\
&\cgres{3}{\lf{\lambda x\lambda y\lambda z.\so{takeflight}(\so{buy}\,z\,\so{ticket}x)\,y\,x}}\\
&\ml{3}{{\tt pre:} $able(x,\so{buy}(z,\so{ticket},x)),flight(y), payable(\so{ticket}),funds(F),have(x,F)$}\\
&\ml{3}{{\tt add:} $at(x,dest(y)),at(y,dest(y)),have(x,\so{ticket}),able(x,z)$}\\
&\ml{3}{{\tt del:} $at(y,orig(y)), at(x,here)$}\\
\cgline{4}{apply}\\ 
\cgres{4}{S\lf{\lambda y\lambda z.\so{takeflight}(\so{buy}\,z\,\so{ticket}\so{john})\,y\,\so{john}}}\\
\ml{4}{{\tt pre:} $able(\so{john},\so{buy}(z,\so{ticket},\so{john})),flight(y), payable(\so{ticket}),funds(F),have(\so{john},F)$}\\
\ml{4}{{\tt add:} $at(\so{john},dest(y)),at(y,dest(y)),have(\so{john},\so{ticket}),able(\so{john},z)$}\\
\ml{4}{{\tt del:} $at(y,orig(y)), at(\so{john},here)$}\\
}
}
\end{center}
\caption{A plan structure for John flying after he himself buys a ticket.}
\label{fig:subsplan}
\end{figure}

\begin{figure}[h!]
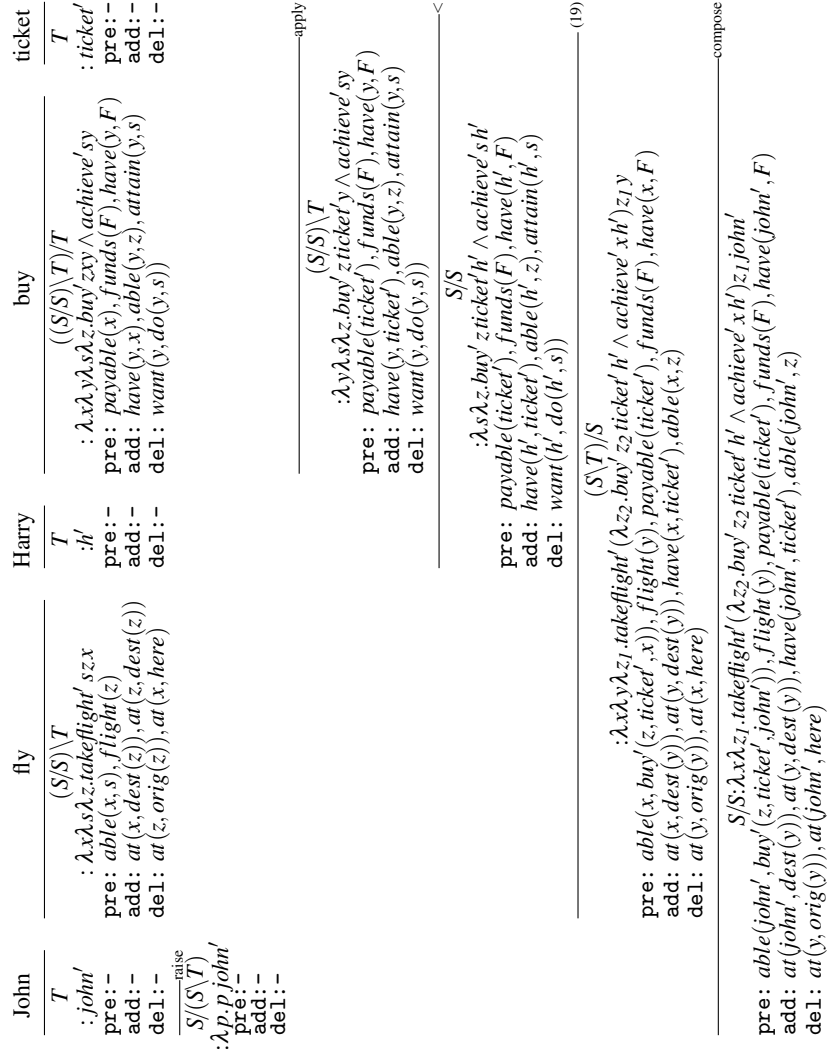

\begin{center}\footnotesize
\rotatebox{90}{
\cgex{5}{John & fly & Harry & buy & ticket\\
\cgul & \cgul & \cgul & \cgul & \cgul \\
\cgf{T} & \cgf{(S\fs S)\bs T} & \cgf{T} & \cgf{((S\fs S)\bs T)\fs T} & \cgf{T}\\
$:\so{john}$ & $:\lambda x\lambda s\lambda z.\so{takeflight}\,s\,z\,x$ & 
\lf{\so{h}} & $:\lambda x\lambda y\lambda s\lambda z.\so{buy}zxy \wedge \so{achieve}sy$ &$:\so{ticket}$\\
\ml{1}{\tt pre:-}  & \ml{1}{{\tt pre:} $able(x,s),flight(z)$} 
& \ml{1}{\tt pre:-} &\ml{1}{{\tt pre:} $payable(x),funds(F),have(y,F)$} &\ml{1}{\tt pre:-}\\
\ml{1}{\tt add:-} &\ml{1}{{\tt add:} $at(x,dest(z)),at(z,dest(z))$}
&\ml{1}{\tt add:-}& \ml{1}{{\tt add:} $have(y,x),able(y,z),attain(y,s)$}&\ml{1}{\tt add:-}\\
\ml{1}{\tt del:-} &\ml{1}{{\tt del:} $at(z,orig(z)), at(x,here)$} 
&\ml{1}{\tt del:-}&\ml{1}{{\tt del:} $want(y,do(y,s))$} &\ml{1}{\tt del:-}\\
\cgline{1}{raise}\\
\cgres{1}{S\fs(S\bs T)}\\
\lf{\lambda p.p\,\,\so{john}}\\
\ml{1}{\tt pre:-}\\
\ml{1}{\tt add:-}\\
\ml{1}{\tt del:-}\\
&& &\cgline{2}{apply}\\ 
&&&\cgres{2}{(S\fs S)\bs T}\\
&&&\cgres{2}{\lf{\lambda y\lambda s\lambda z.\so{buy}\,z\,\so{ticket}y \wedge \so{achieve}sy}}\\
&&&\ml{2}{{\tt pre:} $payable(\so{ticket}),funds(F),have(y,F)$}\\
&&&\ml{2}{{\tt add:} $have(y,\so{ticket}),able(y,z),attain(y,s)$}\\
&& &\ml{2}{{\tt del:} $want(y,do(y,s))$}\\
&&\cgline{3}{\cgba}\\ &&\cgres{3}{S\fs S}\\
&&\cgres{3}{\lf{\lambda s\lambda z.\so{buy}\,z\,\so{ticket}\so{h} \wedge \so{achieve}s\,\so{h}}}\\
&&\ml{3}{{\tt pre:} $payable(\so{ticket}),funds(F),have(\so{h},F)$}\\
&&\ml{3}{{\tt add:} $have(\so{h},\so{ticket}),able(\so{h},z),attain(\so{h},s)$}\\
&&\ml{3}{{\tt del:} $want(\so{h},do(\so{h},s))$}\\
&\cgline{4}{~(\ref{rul:s-like})}\\ 
&\cgres{4}{(S\bs T)\fs S}\\
&\cgres{4}{\lf{\lambda x\lambda y\lambda z_{1}.\so{takeflight}(\lambda z_{2}.\so{buy}\,z_{2}\,\so{ticket}\,\so{h} \wedge \so{achieve}\,x\,\so{h})z_{1}\,y}}\\
&\ml{4}{{\tt pre:} $able(x,\so{buy}(z,\so{ticket},x)),flight(y), payable(\so{ticket}),funds(F),have(x,F)$}\\
&\ml{4}{{\tt add:} $at(x,dest(y)),at(y,dest(y)),have(x,\so{ticket}),able(x,z)$}\\
&\ml{4}{{\tt del:} $at(y,orig(y)), at(x,here)$}\\
\cgline{5}{compose}\\ 
\cgres{5}{\cat{S\fs S}\lf{\lambda x\lambda z_{1}.\so{takeflight}(\lambda z_{2}.\so{buy}\,z_{2}\,\so{ticket}\,\so{h} \wedge \so{achieve}\,x\,\so{h})z_{1}\,\so{john}}}\\
\ml{5}{{\tt pre:} $able(\so{john},\so{buy}(z,\so{ticket},\so{john})),flight(y), payable(\so{ticket}),funds(F),have(\so{john},F)$}\\
\ml{5}{{\tt add:} $at(\so{john},dest(y)),at(y,dest(y)),have(\so{john},\so{ticket}),able(\so{john},z)$}\\
\ml{5}{{\tt del:} $at(y,orig(y)), at(\so{john},here)$}\\
}
}
\end{center}
\caption{A plan structure of John's flying achieving a goal of Harry's buying a ticket for him.}
\label{fig:s-like}
\end{figure}

Now suppose further that in the last variation of interpretation on the same scene, someone else buys the ticket for John to fly, with an expectation (e.g. for John to be away, etc.).
For the overt sequence, and with a different semantics for buying,
the states of affairs will unfold as in Figure~\ref{fig:s-like}. The rule at work is the following (`\,\us\,' means either slash). 
\ex\label{rul:s-like}
\cat{(X\fs Y)\bs Z}\lf{f}~~\cat{Y\us W}\lf{g}~~$\Rightarrow$~~\cat{(X\bs Z)\us W}
\lf{\lambda x\lambda y.fy(gx)}
\xe
We can entertain questions such as 
\emph{Why did  Harry buy the ticket for John?} in a different setting because the result given for the sequence by this rule is \cat{S\fs S}, rather than \cat{S} of Figure~\ref{fig:subsplan}. Notice that we would be questioning an event which has not happened yet (because of `\cat{\fs S}'); it is about expectations inferred from this category of `buy' (other categories are as before). We will come to John's unusual category in Figure~\ref{fig:s-like} soon.

This is yet another structure of organized behavior, which is again not expected from the specified structure of plans in the sequence
but from sequencing itself and its affordances. Next we consider another possible combination.

If the sequence in\xref{ex:pickup2} meets conventional expectation, that is, if it means John takes the flight in a plane rather than buy or use the plane, with the fly-plan having the same semantics as in Figure~\ref{fig:jfbt}, Figure~\ref{fig:subsplan} and Figure~\ref{fig:s-like}, then the sequence altogether can act as an affordance for buying a ticket for the flight, or some similar affordance for boarding a plane. This situation unfolds as in\xref{der:pjf} where plan conditions are omitted for brevity and the projected affordance (means to fly)  is captured in $(hx)$. ({This predicate abstraction \emph{because of sequencing} is not necessary in language, although for example transformationalism appears to consider it to be sufficient, in the form of movement,  at the expense of powerful extragrammatical devices to keep it under control.
See \cite{stee:20} for extensive discussion.)}
\ex\label{ex:pickup2}
plane John fly
\xe

\ex\label{der:pjf}\small
\cgex{3}{plane & John & fly\\
\cgul & \cgul & \cgul\\
\cgf{T}$:\so{plane}$ & \cgf{T}$:\so{john}$ & \cgf{(S\fs S)\bs T}\\
&&$:\lambda x\lambda s\lambda z.\so{takeflight}\,s\,z\,x$\\
& \cgline{1}{raise}\\ & \cgres{1}{(S\fs S)\fs((S\fs S)\bs T)\lf{\lambda p.p\,\so{john}}}\\
& \cgline{2}{apply}\\ & \cgres{2}{S\fs S\lf{\lambda s\lambda z.\so{takeflight}\,s\,z\,\so{john}}}\\[1ex]
\cgline{1}{raise}\\ \cgres{1}{S\fs(S\bs T)\lf{\lambda p.p\,\,\so{plane}}}\\
\cgline{3}{subcompose}\\ \cgres{3}{S\fs(S\bs T)\lf{\lambda h\lambda x.\so{takeflight}(hx)\,\so{plane}\so{john}}}
}\xe

It composes two functions inside another one after abstracting over one of them, which I will call \emph{subcomposition}. Subcomposition's semantics is as follows, which was discovered by \cite{Hoyt:08} in studying linguistic structure, who called it \combd.\footnote{Not to be confused
with Rosenbloom's \citeyear{Rose:50} combinator \combd, which is
equivalent to \combb\combb. Rosenbloom's \combd\,and subcomposition are different: 
$\combb\combb fg$=$\combb(fg)$=$\lambda h\lambda x.fg(hx)$; see \cite{Bozsahin:12} for the differences.}
 I write for the specific sequence in\xref{der:pjf}.
\ex\label{rul:subc}
\cgf{X\fs(Y\bs Z)}$:f$~~~\cgf{Y\fs W}$:g$ $\Rightarrow$ \cgf{X\fs(W\bs Z)}$:\lambda h.f(\lambda x.g(hx))$\hfill(\combd)
\xe
Notice the inner composition of $g$ and $h$ to lead to the argument type of $f$ for \cgf{X}. It symbolizes the affordance of a function ($h$) rather than a term.

Its unfolding is more evident with one less
function abstraction. Below $h$ appears in the sequence, not abstracted over. \cgf{Y\fs W} and \cgf{W\bs Z} can now be clearly seen to be the categories
involved in the composition onto \cat{Y\bs Z}, viz. 
the argument category of $f$.
\ex\label{rul:subcfull}
\cgf{X\fs(Y\bs Z)}$:f$~~~\cgf{Y\fs W}$:g$~~~\cgf{W\bs Z}$:h$ $\Rightarrow$ \cgf{X}$:f(\lambda x.g(hx))$
\xe
It is clearly different from composition and substitution. Because of the fly-plan semantics, and from the fact that in\xref{ex:pickup2} the instrument is made available, it creates the expectation $(hx)$ in\xref{der:pjf}. 
Its plan type, \cgf{S\fs(S\bs T)} is the kind of action we would expect from affordances of objects, for example terms (T),
as we did for John and plane in\xref{der:pjf} and Figure~\ref{fig:s-like}. This is how \cite{Stee:02} uses affordances of Gibson, as abilities of objects
in an environment:
(we can read the result $\lambda p.p\,\so{t}$ as 'whatever \so{t} can do in $p$')
\ex\label{t-raise}
\cgf{T}$:\so{t}$ $\Rightarrow$ \cgf{X\fs(X\bs T)}$:\lambda p.p\,\so{t}$\hfill (type-raising; \combt)\\
\cgf{T}$:\so{t}$ $\Rightarrow$ \cgf{X\bs(X\fs T)}$:\lambda p.p\,\so{t}$
\xe
where \cgf{X\bs T} and \cgf{X\fs T} are restricted to be some function which can take the term \cgf{T} in some plan. Therefore the result of\xref{t-raise} is a second-order function targeting first-order functions like plans. 

This is the lexical view of type-raising, as opposed to syntactic view, where it can be a higher-order function. The lexical view guarantees that it applies once, to a first-order function, whereas the syntactic view can let it apply many times, leading to {\it NP}-completeness \citep{pent:06}, even to undecidability \citep{Hoff:93}.

However, the final affordance in\xref{der:pjf} arises from plane-John-fly, not from a term, and produces the type \cgf{S\fs(S\bs T)} which
one would expect from a term, much like \emph{plane} afforded in\xref{der:pjf}. It shows that anything can be a script, therefore causer of a plan state. 
It creates a script abstraction on the fly. This is going to be one crucial difference from language. In mathematical terms,\xref{t-raise} is Curry and Feys' \citeyear{curryfeys58} \combt, with the definition $\combt\equiv \lambda x\lambda y.yx$, therefore
its unary use is $\combt a\equiv \lambda y.y\,a$.

We shall see that complications that would arise for keeping combinators other than \combb\, and \combs\, in linguistic theory far outweighs
their utility in planning, to be included as part of the joint explanation. (We shall also see that a freely operating \combs\,in language may not be necessary either.)
Their relevance to planning must be considered one way two projective systems with specialized categories can diverge in time, and allows us to be explicit about
what is at stake in generating predicate abstractions in projection.

At this point we might wonder whether there is an end to adding more combinators to project more meaningful sequences locally. \cite{curryfeys58} enumerate all known combinators, which are many,
and we know that two combinators (\combs\,and \combk) are enough to give us any Turing-computable function, which together would clearly be an overshoot, for we know that some undecidable problems such as the Halting Problem of the Turing machine cannot work by finite means to make the sequence interpretable, yet it is definable by \combs\combk. \cite{pian:20} turns this result to advantage,
by prioritizing finitely-computable (i.e. halting) symbolic processes on their run-time, based on combinators for concept formation, in deciding prior probabilities of Bayesian belief update on concepts.
\cite{Bozsahin:12} discusses what is at stake in using \combk\,in linguistics, which would make the interpretation of the sequence not resource-sensitive, that is, not dependent on finite use of all the material in the sequence. Combinations with \combk\,would look like the following, from \cite{Bozsahin:12}:
\ex\label{ex:k}
\cgf{X}$:a$~~~\cgf{Y}$:b$~~~$\Rightarrow$ \cgf{X}$:a$~~~~because \combk$\stackrel{def}{=}\lambda x\lambda y.x$\hfill (\combk)
\xe
where `\cgf{Y}$:b$' resource in the sequence is wasted, because it is eliminated in this combination without possibility of use.
(This will turn out to be intensionally different from ignoring or evading a present resource; see \S\ref{sec:evade}.)

A natural implication of the result to planning is that 
although some elements of the plan may intertwine with other activities
in the world of the agent so that we would not expect all resources in the sequence to be relevant to just one ultimate goal of action, finite use of all resources in planning is expected to be realizable, which \combk\,also violates. 
Mathematical exploration of resource sensitivity is one avenue for understanding the common bond in language and planning.

Before we turn to language for assessing the proposed common representational origin for planning and language, we provide mobile semantics for plan categories we have discussed so far. 

\section{Mobile semantics}\label{sec:mm}\label{sec:mobile}
The kind of plans we have looked at so far  involved one planner only although we have seen examples where other agents can be used by the planner instrumentally, such as someone buying a ticket for someone else to fly.  In this section I show that the dynamic aspect of plans only affects the mobile form, hence they are
uniform across the colloquy of agents, and they must refer to the mobile logical form (MLF) to remain compositional, and to have access to thematic structure in scripts.

\subsection{$\pi$-calculus}\label{sec:pi}
The semantics of $\pi$-calculus is originally designed to handle synchrony in concurrent processes in computer science.
The  formulation of \cite{miln:92a} is summarized below in the simplified notation of \cite{sang:96}.\footnote{It is further simplified to reduce notational clutter. We write the input action $x(y)$ as \piin{x}{y}.
In computer science parlance, a `process' is an agent, for example an executing program, that evolves, that is, changes state. Evolution of a process is denoted by '$\Rightarrow$' in $\pi$-calculus. When labelled, as in $P\stackrel{\alpha}{\Rightarrow} Q$, \emph{action} $\alpha$ is said
to evolve $P$ to $Q$.} In Milner's (\citeyear{miln:92}) terminology, $\lambda$-calculus communicates arguments to sequential functions, and $\pi$-calculus does the same for mobile processes.
\ex\label{def:picalc}
Processes $P ::= \piin{x}{y} \mid \piou{x}{y} \mid \pisi \mid (\pisu{P_1}{P_2}) \mid \pina{x}{P} \mid (\pipa{P_1}{P_2}) \mid\, \pire{P}\mid\,\pize$
\xe
Processes ($P$s) are defined recursively. The simplicity and expressiveness of the calculus come from its treatment of \emph{names} as the only objects
of mobility,  as both channels of information, and also information itself. This property
is the main expedient in harnessing it for actions of the mind without infinite regress or stratification.

The first three cases of the notation above are known as prefixes, or \emph{guards}.
It is common to write a dot before the guarded.
They are actions of processes (i.e. prefixes to $P$ above). 
Processes are assumed to be  in need of various forms of information exchange:
\piin{x}{y} is  inputting in channel $x$ of the name $y$. \piou{x}{y} is outputting in channel $x$ of name $y$. \pisi\, is silent (noncommunicative) action. 

For example, \piin{x}{y}$.P$ would be process $P$ waiting for input $y$ in channel $x$ to proceed; \piou{x}{y}$.P$ would be  process $P$ emitting in channel $x$ the name $y$ before it proceeds. $\pisi.P$ is easier to describe in 'evolves' notation:
$\pisi.P \Rightarrow Q$ means $P$ evolves into $Q$ without interaction with the environment. \pisi\,is possible too, meaning `silent within same agent'.

The sum process, +, called \emph{choice}, behaves like one or the other, such as $P_1$ or $P_2$ in\xref{def:picalc}. The naming process, $\nu$, called \emph{restriction},
creates a new name. \pina{x}{P} behaves like $P$ except that there is no communication action in the newly created name $x$. Concurrent processes are denoted by 
\pipa{P_1}{P_2}.  The replication process, !, creates as many copies of $P$ as needed in \pire{P}. The null process, \pize, completes execution and terminates. (Therefore it is different than the `silent' process.)
A process is essentially any function which is possibly guarded by prefixes, as recursively captured in\xref{def:picalc}.

We shall assume that if the process is not guarded and simple,
that is, if the prefix is not a $\pi$-calculus operation such as a guard,
choice, restriction, replication, or null process, then we are down to $\lambda$-calculus, which is sequential communication of arguments to their functions. 
The dots of $\pi/\lambda$-calculi are disambiguated by the body on the right. It reduces notational clutter.

\subsection{Semantics of mobility}\label{sec:mobsem}
The main semantics that concerns us here is the standard reduction rule of $\pi$-calculus which allows concurrent computation to proceed:
\ex\label{def:pisem}
(\piou{x}{z}$.P$ \pipa\,\,\piin{x}{y}$.Q$) $\Rightarrow$ ($P$ \pipa\,\, $Q\lbr z/y\rbr$)\\
 where $Q\lbr z/y\rbr$ is substituting the free occurrences of $y$ in $Q$ with free name $z$.
\xe
It means that processes $P$ and $Q$, whose timing and duration are not known to each other, because of \pipa{},
meet by $P$ supplying $Q$ with an input in a channel that is known to each other (perhaps agreed in advance, or by chance), namely $x$. What is supplied, $z$, and what is expected, $y$, are local. Thereafter $P$ and $Q$ go their own ways, because of \pipa{}.

Additional semantics of mobility can help us understand what takes place in $\pi$-calculus:
\ex\label{def:pisem2}
if $P \Rightarrow Q$ then $\pipa{P}{R} \Rightarrow \pipa{Q}{R}$\hfill(concurrency does not impede computation)\\
if $P \Rightarrow Q$ then $\pina{x}P \Rightarrow \pina{x}Q$\hfill(restriction does not impede computation)\\
if $P\!\equiv\!P\,' \wedge\,\, Q\!\equiv\!Q\,' \wedge\,\, P\!\Rightarrow\!Q$ then $P\,'\Rightarrow Q\,'$ \hfill(eqv. processes reduce the same)\\
$\pire P\equiv \pipa{P}{\pire P}$\hfill(replication is concurrent and recurrent)
\xe

As pointed out by \cite{boud:92}, the original $\pi$-calculus is synchronous. The semantics of\xref{def:pisem} is such that
the ``sending process'' $P$ is not allowed to continue before its message is received by some process, in this case $Q$.
This is a requirement because $Q$ must be able to do the substitution in (\ref{def:pisem}).

Boudol introduces asynchrony to the calculus by providing a subset of\xref{def:picalc} where all sending processes are terminal.
That is, if the prefix is \piou{x}{y}, then the process can only be \pize. Because it is guaranteed to do nothing after its sends, we achieve asynchrony where messages float in cyberspace looking for a receiver. (Boudol shows an abstract machine for that.)
It is a subset of $\pi$-calculus in which the receiver knows that the sender has terminated. If all agents work with this assumption,
they  adopt asynchrony.
\cite{pala:03} proved that such a subset is less expressive than the full calculus.

Now let us put asynchronous $\pi$-calculus to work in mindful action. 

\subsection{Mobility, affordances and namespaces}

Consider the following example, which will turn out to be a mobile version of finding a means to fly.\footnote{It is clear
that in the  process headed by \so{fly}, `$\lambda z.$' is not a guard but a $\lambda$-term. We will continue to use the $\lambda$-calculus
dot notation when no confusion arises.}
\ex\label{ex:pi1}
$\pina{x}{(\pipa{\piou{x}{a}.\pize}{\piin{x}{y}.\piou{y}{x}.\lambda z.\so{fly}\so{plane}yz}.\pize)}\pipa\piin{a}{t}.\piou{t}{t}.\pize$
\xe
The first two concurrent processes in the parenthesis
share a channel, $x$. For example, it can be thought of as obtaining and sending a ticket to the flyer.
The third one (outside the parenthesis) will share a channel as the processes unfold, because of \piou{x}{a}. 
The plan proceeds depending on which channel is made available:
\pex[labeltype=numeric,nopreamble]\label{ex:pi1der}
\a $\pina{x}{(\pipa{\piou{x}{a}.\pize}{\piin{x}{y}.\piou{y}{x}.\piin{x}{y}.\lambda z.\so{takeflight}\so{plane}yz}.\pize)}\pipa\piin{a}{t}.\piou{t}{t}.\pize \Rightarrow$\hfill $\lbr a/y\rbr$ on channel $x$
\a $\pina{x}{(\pipa{\pize}{\piou{a}{x}.\piin{x}{y}.\lambda z.\so{takeflight}\so{plane}yz}.\pize)}\pipa\piin{a}{t}.\piou{t}{t}.\pize\Rightarrow$\hfill $\lbr x/t\rbr$ on channel $a$
\a \pipa{$\pina{x}{(\pipa{\pize}{\piin{x}{y}.\lambda z.\so{takeflight}\so{plane}yz}.\pize)}}{\,\piou{x}{x}.\pize}\Rightarrow$\hfill $\lbr x/y\rbr$ on channel $x$
\a $\pina{x}{(\pipa{\pipa{\pize}{\lambda z.\so{takeflight}\so{plane}xz.\pize})}{\pize}}$
\xe
For a while we will not write \pize\,if it is a suffix of a process. However, it will turn out to be crucial for language. This has been already hinted
by adopting Boudol semantics, by concentrating on processes in which the receiver knows the sender has terminated.

Notice that, going from state 1 to 2, the inner name $x$ of process 2 is not substituted. The same is true of  $y$ of inner $xy$.
Going from state 2 to 3 is made possible by $a$ becoming a common channel. Going from state 3 to 4 is made possible by  channel $x$ revealing itself again in due course. Since a local name ($x$) had been output going from state 2 to state 3, its scope has been extended now to cover all three processes. This is called \emph{scope extrusion} by \cite{miln:92a}.
The first and third processes can terminate on their own, the first one after state 2, and the third after state 4. The second one terminates when the flight is over. 

Apart from input-output dependencies between the processes, nothing needs to be said about their sequencing, since they go about their computations in their own ways. Roughly, the agent who wants to take a flight, $z$,
receives an affordance for it first (perhaps a ticket, where the ticket provider and receiver work concurrently), then seeks 
another facility to make flight possible (e.g. boarding, whose agent similarly operates concurrently 
with the flyer except where they meet), and the flyer takes the flight with the restriction that allowed the plan in the first place (e.g. ticket). None of that is fixed in advance, except someone's desire to get on a plane. 

There is only one mobile logical form
in the processes of\xref{ex:pi1} and\xref{ex:pi1der}.
This is quite different that earlier conception of scripts in AI/cognitive psychology with preconceived roles about means of doing things such as \cite{scha:77}. It also differs from similarly inspired approaches to lexicalized reasoning such as \cite{geib:15,geib:09-ai,McMi:06}, where categories bear roles. Indirect correspondence of a few primitive categories such as \cgf{S} (state) and \cgf{T}
(term) with the logical form, and \emph{its} compositional but indirect relation with the mobile form, are the key features of this way of thinking, as we have done in\xref{ex:pickup}. We will add to it the mobile aspects shortly.

The phrasing of steps in\xref{ex:pi1der} on the right might appear to suggest that channels are Gibsonian (\citeyear{Gibson:77})
affordances created by substituted values. That would be an informal way to interpret affordances. They are in fact second-order functions. Unlike communication needs being served by message passing, affordances in\xref{t-raise} do not know what the needs are, because we don't know in advance
what \so{t} can do in $p$. Affordances  are categorial projections
of functionality to later
stages of combination. As we shall see, these mobile needs are also different than the terms of the Frame Problem's precondition, add, and delete lists.

\subsection{Synchrony, asynchrony and collaboration}

Asynchrony is already present in\xref{ex:pi1} and\xref{ex:pi1der}. In the running example interpretation
of the states of affairs, the ticket provider and the boarder can go their own ways in meeting the flyer while they send and receive their input and output. In computer science terminology, the first process of\xref{ex:pi1der} is ``nonblocking'' after state 1, and the third process after state 3.

We provide another example which makes crucial use of synchrony along with asynchrony. The problem is distinguishing dancing in the rain from scurrying in the rain, which \cite{sear:90} uses to argue for the necessity of having i-intentions (`i' is for internal.
Searle uses this to argue that humans have different mental representations for dancing and scurrying in the rain.) 

From the viewpoint of an external observer, the overt actions may look exactly the same, people moving about in rain, and only having a collaborative plan can distinguish intentional action. We develop two mobile semantics as proxies for the distinction in mental representation.\xreff{ex:dance}{a} is planned action, (b) is just action. 
\pex\label{ex:dance}
\a $\pire_i{\pina{x_i}\pire{(\pipa{\piou{x_i}{y}}{\piin{x_i}{x_i}}.\lambda z.\so{move}x_iz )}}$\hfill dance
\a $\pina{x}\pire(\pipa{\piou{x}{y}.\so{shelter}x.\pize}{\piin{x}{x}.\lambda z.\so{move}xz})$\hfill scurry
\xe
The main overt action is same: \emph{move}.

In (a) every coordinated move creates a name space. The index of the first \pire{} represents epochs in the dance.\footnote{The semantics of indexing is the following:
$\pire_i{P}\equiv \pipa{P_{\lbr i\rbr}}{\pire_i{P}}$, where $P_{\lbr i\rbr}$ means process $P$ whose namespace
can use the fresh index $i$.}
 The number of dancers, which is the second  \pire{} in (a), is left unspecified. Let us assume it is $m$. For fair comparison we shall assume that the number of dancers ($m$) and the number of people scurrying ($n$) are the same, $n$=$m$. Scenario (a) unfolds as follows:
\ex\label{ex:danceder}
	$\pina{x_1}((\pipa{\piou{x_1}{y}}{\piin{x_1}{x_1}.\lambda z.\so{move}x_1z})
\cdots\pior_n(\pipa{\piou{x_1}{y}}{\piin{x_1}{x_1}.\lambda z.\so{move}x_1z}))$\\
\hspace*{9em}$\vdots$\\
$\pina{x_m}((\pipa{\piou{x_m}{y}}{\piin{x_m}{x_m}.\lambda z.\so{move}x_mz})
\cdots\pior_n(\pipa{\piou{x_m}{y}}{\piin{x_m}{x_m}.\lambda z.\so{move}x_mz}))$\\
\xe
As namespaces \pina{x_i}{} make themselves available, coordination within an epoch or stage
is realized by nonblocking message passing (\piou{x_i}{y}), and it monitors epoch $i$ locally, staging $x_i$ moves by several agents of \so{move}, symbolized by $z$. There is no central guard to go to next epoch. We assume for exposition that all staging is simultaneously available but acted in agreed sequence $1\cdots m$. This can of course be manipulated differently.

Scenario\xreff{ex:dance}{b}, accidental similarity, works as follows. An available shelter is spotted by independent agents, and they move toward that and stay there (\pize). Even if they enact exactly the same sequence of \so{move}, the representational bases of the moves are different than\xreff{ex:danceder}{a} at Marr's level-2, therefore the mental correlate at level-3 may be different. Mobility here is a level-1 concept, synchrony and asynchrony are its subconcepts, 
engendering the difference for example
in\xref{ex:dance}.

\subsection{Namespaces and adversaries}\label{sec:evade}

$\pi$-calculus is expressive enough to model adversarial planning without additional mechanisms.
In the following scenario (a), $P_1$ can be said to be adversary to $P_2$ because, although it communicates, it passes on nothing that $P_2$ can use, at the same time using $P_2$'s channel ($y$) to do something.\footnote{The fact that mobile semantics must rely on someone's own resources \emph{and} that of others, created or imagined by others, and do so by finite enumeration of   intentional and nonintentional properties that can be communicated, appears
to be a prerequisite to what \cite{bull:07,bull:09} called an \emph{ontological commitment} to agent tracking. This commitment is physical
in \cite{pask:68,pask:76}.}
 In (b) $P_1$ is evasive; it passes on information that $P_2$ cannot make use of as something coming from $P_1$.
In (c) $P_1$ is deceptive; it passes on  information that itself does not make use of. 
Example (d) is not nondeterminism. It is not evasive or deceptive, but not revealing either, where 
an agent either dances or scurries.
\pex\label{ex:adv}
\a$\pipa{\piou{x}{(\lambda x.x)}.\piin{q}{y}.P_1y}{\piin{x}{y}.\piou{q}{r}.P_2ry}$
\a$\pipa{\piou{x}{(\lambda x.x)}.P_1}{\piin{x}{y}.P_2y}$
\a$\pipa{\piou{x}{y}.P_1x}{\piin{x}{y}.P_2y}$
\a $\pire_i{\pina{x_i}\pire{(\pipa{\piou{x_i}{y}}{\piin{x_i}{x_i}}.\lambda z.\so{move}x_iz )}}
+\pina{x}\pire(\pipa{\piou{x}{y}.\so{shelter}x.\pize}{\piin{x}{x}.\lambda z.\so{move}xz})$
\xe
Therefore the range of collaborative plans that can be captured at level-2 can target real level-3 situations.

\subsection{Mobile semantics and the Theory of Mind}
The Theory of Mind (ToM), awareness of the difference of self's psychological spaces about a state of affairs from that of others, and understanding its consequences, is a developmental process \citep{wimm:83}. Human infants are assumed to have developed a competence of it  around  the age of five, much later than syntactic-morphological performance, which begins to show signs of competence by the age of two. 

ToM was first proposed
for understanding the complex behavior of chimpanzees and their differences from that of  humans \citep{prem:78}. This way of thinking and its experimental methods, in fact the very presence of ToM in nonhumans, has been questioned \citep{penn:07}, in some cases by the same researchers changing their minds by further experimentation,
from ascribing very little ToM to chimpanzees \citep{toma:97,toma:03}, then to chimpanzees having ToM but lacking belief-desire psychology \citep{toma:08}.

Mobile Semantics ought to say something about ToM, or whatever its functional equivalent is, sometimes called \emph{intersubjectivity}: 
``Far oftener than any of us are aware, humans intuit the mental experiences of other people, and---the really interesting thing---care about
having other people share theirs.'' \citealt{hrdy:11}:2. I will provide no further than a first-degree representational support for beliefs about beliefs using mobile semantics, rather than argue for 
meta-representations (cf. \cite{denn:78,pyly:78}).

To take a modern-day typical experimental setting for children's ToM, assume that Sally and Ann are in the same room. There is a covered basket and a box, say represented as \so{basket} and \so{box}. Sally has a ball (\so{ball}). She puts it into her basket (\so{in}), covers it, and goes out for a walk. Ann takes the ball out and puts it in her box. Sally comes back, and wants to take the ball. `Where would Sally look for the ball' is the ToM question to the subject observing the interaction. We can think of the beliefs and events unfolding as follows:

\pex
\a\begin{minipage}[t]{2.3cm}\scriptsize
After Sally puts the ball in the basket,
before she goes out, Ann observing, subject observing them:
\end{minipage}
{\small\begin{tabular}[t]{l@{}l@{}l}
& $\pina{x_{1}}($\\
Sally's MLF: &  &$\piou{x_{1}}{y_{1}}.\so{in}(y_{1},\so{basket}) \wedge y_{1}=\so{ball} \pipa$\\
Ann's MLF: & & $\piin{x_{1}}{z_{1}}.\so{in}(z_{1},\so{basket}) \wedge z_{1}=\so{ball} \pipa$\\
Subject's MLF: & &$\piin{x_{1}}{w_{1}}.\so{in}(w_{1},\so{basket}) \wedge w_{1}=\so{ball}$\\
& ~~~~~~$)$
\end{tabular}}

\a\begin{minipage}[t]{2.3cm}\scriptsize
After Ann changes ball's location, Sally comes back in, and subject is asked
the ToM question:
\end{minipage}
{\small\begin{tabular}[t]{l@{}l@{}l}
& $\pina{x_{2}}{}\pina{x_3}{}($\\
Sally's MLF: &  &$\pisi.\so{in}(\so{ball},\so{basket})\,\pipa\,x_{2}z_2.\pize$\\
Ann's MLF: & & $\pisi.\so{in}(\so{ball},\so{box})\,\pipa\,x_{3}z_3.\pize$\\
Subject's MLF: & & $\piou{x_2}{z_2}.\piou{x_3}{z_3}.\so{in}(\so{ball},\so{box}) \pipa$\\
&&$(z_2\!=\!\so{in}(\so{ball},\so{basket})  +$\\
&&~~$z_3\!=\!\so{in}(\so{ball},\so{box}))$\\
& ~~~~~~~~~~~~$)$\\
\end{tabular}}
\xe
In (a), Sally does not overtly communicate (assuming \pina{}\, is the subject's perspective of what Sally and Ann do), and \piou{x_{1}}{} establishes
observable  common background by joint attention  (\piin{x_{1}}{}) on basket and ball by Sally. The timing of attention is not metrically constrained.
According to\xref{def:pisem2}, once the process (a) begins to evolve, it will have established the common ground by names.

In (b), where \pina{x_2x_3}{} is  also assumed to be set up by the subject, presumably from having an interest about actions of both Sally and Ann, Ann can be seen not to report the box's contents to Sally, and this is externally observable by the subject,  from having both \piou{x_{2}}{} and \piou{x_3}{}, i.e. no communication from others, no communication to others---$z_2$ and $z_3$  of Sally and Ann are unused and their processes are terminal; but, recordkeeping in self's namespace about what Sally and Ann did,
respectively \piou{x_2}\, and \piou{x_3}. 

The cognitive development seems to be that, knowing that the ball is in the box, independent of what Sally or Ann thinks about it (i.e. without dependence on \emph{their} names for this fact), the subject
becomes aware of two views on the states of affairs, not one, therefore a choice (+) is to be made by the subject (see\xref{def:picalc} for `+' process), apparently randomly 
before reaching the age of  five, and no  choice before the age of three or four \citep{wimm:83}.

With mobile semantics, beliefs about others' beliefs can translate to cross-referential competence with awareness of choice. Beliefs about beliefs are first-class citizens of mobile logical forms.

\subsection{Indefinite plan extension and partial observability}\label{sec:cut-tree}
As demonstrated by \cite{leve:05,hu:11}, temporal extension of a plan may need loops, where same finite sequence of actions and observations may lead to possibly indefinitely extended planning domains, such as felling a tree
by repeated chops. The agent of such plans would be limited by his observations,
and partial observability of the plan cannot be eliminated algorithmically from such computable
infinite domains \citep{degi:16}. 

This is quite unlike the world which STRIPS-like planners tend to model because not only the agent is limited but
the partial results achieved are dynamic, rather than propositional
and time-invariant.

It can be given  mobile semantics
with time-dependent channels to combine these two aspects. Partial observability creates conditional processes, and time extension is achieved by replication. We assume that the primitives $look, chop, store$ used by \cite{degi:16} for the tree felling problem, which respectively looks to see if the tree is up or down, chops the tree with the axe, and stores the axe. It has the following definitions where $cond\,x_1x_2x_3$ is shorthand for
``if $x_1$ is true, do $x_2$, otherwise do $x_3$''. As usual we write
the least oblique argument, which is usually the agent, furthermost from the predicate (because it is least dependent on it).
\ex\label{ex:treefellingops}
$\lambda x\lambda y\lambda z.\so{look}\,xyz$=$cond\,(up\,x)yz$\\
$\lambda x.up\,x$= true if $x$ is up.\\
$\lambda x\lambda y.\so{chop}\,xy$= $y$ chops $x$.\\
$\lambda x.\so{store}\,x$= $x$ is stored.
\xe
Tree-felling problem is now definable with mobile meanings of
indefinite number of stages as follows,
where the intended interpretation is obtained with $x$=axe, $y$=tree,
$z$=a stage of chopping.
\ex\label{ex:tfprob}
$\pina{x}\pina{y}(\pire\pipa{\pina{z}(cond\,(up\,y)(\piou{z}{y}.\pize)(\piou{x}x.\pize)}
{\pipa{zy.\so{chop}\,yx}{xx.\so{store}\,x}}))$
\xe
Compare this solution with the finitely and definitely extended version below, where time extension
is enumerated with $y_n$=epochs of chopping, and $i=0\cdots n$. Unlike collaborative plans we have seen earlier for dancing, identification of epochs in such plans is not necessary. 
\ex\label{ex:tfprobfin}
$\pina{x}\pina{y}(\pire_i\pipa{\pina{y_i}(cond\,(up\,y)(\piou{y_{i+1}}{y}.\pize)(\piou{x}x.\pize)}
{\pipa{y_{i+1}y.\so{chop}\,yx}{xx.\so{store}\,x}}))$
\xe
Definite extension by \pire$_i$\,and indefinite extension by \pire\ differ
only intensionally, with an understanding that stages in between have been made part of the plan in advance in the intensional version, and perhaps referred to by the participants. Otherwise the internal semantics of looping is extensionally equivalent because dependencies among the channels can provide the same behavior. Notice the same functional dependency achieved by name $z$ in\xref{ex:tfprob}. Computationally, indefinite extension requires lazy evaluation of \pire\,(that is, evaluate only when needed), which shows that they may be extensionally equivalent but they are not equally expressive.

\subsection{Inherent ambiguity of deliberative planning}\label{sec:lewis}
One final note about the mobile semantics of deliberation. It tells us more about the nature of mobile logical forms, in planning and in language.

Calling a plan $P$ \emph{deliberate} seems to be equivalent to typing it as 
$\pina{x}{(\pisu{P_1\{x\}}{P_2\{x\}})}$ in terms of mobile semantics of\xref{def:picalc}.\footnote{And here I follow
the apparently nonKantian tradition of trying to explore a goal-oriented system
in as much mechanistic terms as possible; see also 
\cite{juar:85,deac:12,gins:19}. `Mechanics' here is to be understood
as prioritizing the \emph{how} question, beginning the exploration with it,
which is at the heart of starting at Marr's level-2, rather than some specific
substrate; see \cite{becht:94} for support and also criticism of 
the levels approach.} 
$P_1$ and $P_2$ are intended here to be alternatives in $P$, and $P\{x\}$ stands for use of a set determined by $x$ in $P$, for example, a
list of participants. We may plan to act on $P$, say as $P_1$, \emph{and} presuppose $P$ as $P_2$, not to act on it but to do performatives. Which one prevails depends on the unfolding states of affairs, both of which  use $x$.  \cite{lewi:79} describes the intensional difference of  contemplating an action versus holding a presupposition about it from the perspective of philosophy:

\begin{quote}
So good is the parallel between plan and presupposition that we might well ask if our plan simply \emph{is} part of what we presuppose. Call it that if you like, but there is a distinction to be made. We might take for granted, or purport to take for granted, that our plan will be carried out. Then we would both plan and presuppose that we are going to steal the plutonium [from a nuclear plant]. But we might not. We might be making our plan not in order to carry it out, but rather in order to show that the plant needs better security. Then plan and presupposition might well conflict. We plan to steal the plutonium, all the while presupposing that we will not. And indeed our planning may be interspersed with commentary that requires presuppositions contradicting the plan. “Then I'll shoot the guard (I'm glad I won't really do that) while you smash the floodlights.” Unless we distinguish plan from presupposition (or distinguish two levels of presupposition) we must think of presuppositions as constantly disappearing and reappearing throughout such a conversation.

The distinction between plan and presupposition is not the distinction between what we purport to take for granted and what we really do. While planning that we will steal the plutonium and presupposing that we will
not, we might take for granted neither that we will nor that we won't. Each of us might secretly hope to recruit the other to the terrorist cause and carry out the plan after all.\hfill\citealt{lewi:79}:357
\end{quote}

It seems that what Lewis calls presuppositions and plans must have thematic roles, for example thief and reward for his plan, 
and inspector and customer (e.g. government) for the presupposition.
Therefore a shared namespace is established, which will project different
thematic roles to things in $\{x\}$. The second paragraph in the quote implies that these $P$s do not treat each other as guards, therefore no communicative action between them is necessary. We can consider each deliberative plan say $P=(P_1,P_2)$ to be inherently typed as $\pina{x}{(\pisu{P_1\{x\}}{P_2\{x\}})}$.

In language, such states of affairs manifest themselves in what is captured by Alternative Semantics of \cite{Root:85,Root:92}, where meaning contrast between two expressions of \emph{same lexical material} (i.e. overt identity) is derived from \emph{two} logical forms which are related to each other by the same namespace.

\subsection{Synopsis: namespaces and grounded planning strategies}
Once we know how we can interpret mobile objects, using $\pi$-calculus for concurrency and a very restricted $\lambda$-calculus for logical forms of sequential functions, there seems to be no need for separate mechanisms of forward reasoning, backward reasoning, affordances versus sequence/plan composition, synchronous versus asynchronous collaborative planning, instrumental planning, multi-agent planning, indefinite versus enumerable extension in time or space, adversarial planning, or planning versus presupposition. All of these aspects can be ``lexicalized'' (i.e. grounded) from the perspective of an agent, representing
a personal view of the (multi-agent) world as strategies \citep{find:12}. 

From the perspective
of society of agents, the world view as it changes can be seen as the one with collective mobilized semantics 
of its members. Logical forms of Section~\ref{sec:plans}
can now be replaced with mobile logical forms (MLFs) of this section, where the ground
cases of communication are simply lambda terms of actions 
with their own precondition, add list and delete lists. Within those lambda terms there is only act, and no communicative act.
It is a society of mindful agents \citep{simo:69}, rather
than a society that builds a mind from mindless agents \citep{mins:88}, because the agents act with awareness of the semantic significance of representational difference.
The idea goes back to \citeauthor{huss:00}; it predates representationalism of the kind we see in cognitive science:
\begin{quote}
Clearly we may say that if presentations, expressible thoughts of any sort whatever, are to have their faithful reflections in the sphere of meaning-intentions, then there must be a semantic form which corresponds to each presentational form. This is in fact an \emph{a priori} truth. And if the verbal resources of language are to be a faithful mirror of all meanings possible \emph{a priori}, then language must have grammatical forms at its disposal which give distinct expression, i.e. sensibly distinct symbolization, to all distinguishable meaning-forms.\\
\hspace*{10em}\emph{Logical Investigations} vol.II: 55 \cite{huss:00}
\end{quote}

Accidental and intentional behavior can be shown to arise from different intensions, without having to spin a tale about what intentions really are and where they come from.\footnote{We can say that they come from subjects. Then explaining the subject as someone or something capable of  subjective experience does not provide much of an advance on the problem  \citep{denn:95}, even if we assume that it is not circular or vitalist. We have
yet to see a convincing proposal about how subjective experience as a presumed ``measurable'' is measured.}
In emerging planning theory parlance, the latter are deliberative pieces of information. The prerequisite representational support for them must involve some way of dealing with
change and no change, and sharing of psychological spaces (not psychological states---channels are exchanged names), online and offline. The last aspect takes us to language.

\section{Language}\label{sec:language}
There is a tendency among nonlinguists to either take language, in  one extreme, as some specialization of a general-purpose conceptual capacity, or some specialization of universal symbolic logic in the other. 

Most linguists would disagree. To be sure, the assumption of
language as ``the releasing agent of symbolic thought; an invention'' 
(\citealt{tatt:16}:164), or product of general associative learning \citep{lieb:16}, are not unheard of, especially in functional linguistics, for example
\cite{langacker87,fauc:94}. From the logic side, it ranges from
Vienna Circle's perfect logic of nature waiting to be discovered
by imperfect creatures (\citealt{carn:67}:2), to assuming no difference
between formal systems and natural language \citep{Mont:73}, which forms
the basis of some linguistic theory, e.g. \cite{Part:73b,Dowt:82}.

The search for an alternative linguistic theory would not change the fact that linguistics is a natural science, not
because of some purported single biological event causing ``merge''' therefore syntax in humans \citep{bolh:14}, or
some cognitive human core for linguistic concepts \citep{langacker87,Jack:97},
but because of the stable domain of observing children. The striking crosslinguistic uniformity of timeline of language acquisition despite intelligence quotients, cultural, economic, educational and local differences, which we have known since \cite{lenn:67}, 
and what happens when children are deprived of acquiring language in action in the critical period \citep[see e.g.][]{from:74},
demand a natural inquiry.

 There seems to be unexpected gaps in linguistic data that defy the oversimplistic scene of an omnipotent inference/concept engine let loose on verbal expressions. Seven examples of this kind are enough to start off the discussion: 
\pex[labeltype=roman,pexcnt=1,labelwidth=2em]\label{ex:gaps}
\a There is no language in which we can extract (relativize) more than one argument out of an embedded clause, although there are varying
intralinguistic and crosslinguistic restrictions on which arguments can relativize.
\a There is no language in which complement-taking verbs take more than one clausal argument, something which is semantically a proposition.  They can take many such things as adjuncts. 
\a There is no language in which the syntactic subject can be a referentially clause-internally-dependent reflexive (in the true sense of the reflexive).
\a There is no language in which we can extract a verb out of an embedded clause, although we can extract all kinds of nonverbs.
\a  There is no language in which a verb root
is a full proposition, rather than predicate. (In pro-drop languages  the verb stem might be full proposition, when
the ``dropped'' argument is \emph{morphologically} available on the verb. Topic-drop is a different phenomenon.)
\a  There is no language which is not structure-dependent. 
\a There is
no language which exhibits
hierarchical dependencies of the kind 3124 and 2413 that we saw in\xref{ex:sts}. 
\xe

There is a longer list, which might get more technical depending on one's favorite linguistic theory, but these will suffice for our purposes as far as syntax-semantics and sequencing are concerned. The linguist's claim is that there has been no such languages, and there will never be even if we could magically fast-forward or fast-backward the evolution of the language-forming capacity, hence the need for a linguistic theory to explain the gaps.\footnote{An analysis of one striking
historical example for the gap in\xreff{ex:gaps}{i} which competes with modern analyses and reaches the same result is the study of old Hittite by
\cite{Bach:78}.}

Case (vii) has already been touched upon, around discussion of\xref{ex:sts}.
Case (vi) has been a Chomskyan classic, sometimes considered
to be the cornerstone of his syntactocentrism, for example arguing for 
structure-dependence of yes/no question inversion in English.\footnote{Case (iii) is Chomsky too (and Haj Ross), that is, syntactically argued, but, like
others, semantic in nature.} They are the ones I will concentrate on (and leave the rest to an upcoming book) because they are brought to focus by categorial grammar, which also tries to explain structure-dependence in (vi) rather than assume it as Chomskyans do \citep{Bozsahin:12}. Unsurprisingly, semantics is brought in to the explanation. More to the point of the current article, they are about sequences humans find connected and interpretable.

\subsection{Neither syntax nor semantics suffices for understanding linguistic structure}
The following examples manifest different semantics depending on the verb used, although the  surface syntactic structure (intonation, grouping, phrasing) is the same for all the verbs listed.
\ex\label{ex:samesyn}
Mary persuaded/promised/expected/wanted John \lbr to study\rbr.
\xe
In `persuade', John studies; in `promise' Mary does. For `expect' and `want', John is only the subject of studying and not an argument of these verbs. 
For example, Mary does not expect John, or want him.
(These are respectively
known as object-control, subject-control/raising-to-object, and exceptional case-marking verbs.) 

They are bounded dependencies, where the verb specifies
the role of its argument  in the infinitival clause in brackets, if any. In other bounded dependencies, such as passive, where the governor likewise uniquely identifies the roles of the particular argument it is interested in, the telicity of the verb makes a difference although the surface syntactic structure is considered acceptable across the board  but  semantically odd in some cases (shown with $\sharp$):
\ex
John was persuaded/expected/$\sharp$promised/$\sharp$wanted \lbr to study\rbr\,(by Mary).
\xe

For the same verbs participating in unbounded constructions, such as relativization, (a--b) below, coordination (c--d), and topicalization (e), where the dependencies between arguments
can cross clausal boundaries indefinitely without being an argument of any of the intervening verbs, we see the same  surface syntactic structure for all the verbs in question.
\pex\label{ex:samesyn2}
\a I met the man that \lbr Harry thought \lbr Jenny claimed \lbr Mary\\ \hspace*{1em}persuaded/promised/expected/wanted to study\rbr\,\rbr\,\rbr.
\a The man that \lbr Harry thought \lbr Jenny claimed \lbr Mary\\ \hspace*{1em}persuaded/promised/expected/wanted to study\rbr\,\rbr\,\rbr\,walks.
\a \lbr Mary asked\rbr\,and \lbr Jenny \lbr believes Larry\\ \hspace*{1em}
\lbr persuaded/promised/expected/wanted\rbr\,\rbr\,\rbr\, John to study.
\a Mary persuaded/promised/expected/wanted \lbr John \lbr to attempt \lbr to leave\rbr\,\rbr\,\rbr\\ \hspace*{1em} and \lbr Jenny \lbr to try \lbr to arrive\rbr\,\rbr\,\rbr 
\a This man, Jenny \lbr knows\\ \hspace*{1em}\lbr Mary persuaded/promised/expected/wanted to study\rbr\,\rbr     
\xe
One important problem then is how we can acquire the semantic distinctions of these verbs if syntax suffices for semantics, because the surface syntactic structures engendered by these verbs are uniform in bounded and unbounded constructions, and presumably, this is the data we are exposed to in acquiring language. One answer, emanating from \cite{Chom:70a}, is that their internally represented syntactic structures are different. Another answer, which CCG follows,
is that surface structure \emph{is} syntactic structure, but their logical command relations differ, which means such command relations must be represented as part of the elements, in this case the verbs, because we know that children can acquire  their distinction.\footnote{`Promise'-class
is notoriously difficult for the child, crosslinguistically. However, they are acquired early \citep{sher:93}, but later than other verbs, around the age
of four years to five and a half \citep{hyam:17}.} However, both theories agree that
the computational process that drives the surface structure is one and syntactic,\footnote{Recall once again the famous Chomsky example \emph{colorless green ideas sleep furiously}, which is considered grammatical but senseless.
The idea of single computation sets Chomskyan transformationalism and CCG apart 
from multi-structural multi-computational proposals in linguistics, such as
HPSG \citep{PandS:94}, LFG \citep{Bres:82c}, autolexicon \citep{Sadock:91}, and ``tiers'' approaches \citep{Jack:02}. The additional burden on latter theories is to explain
how such multiple structures meet if they are computed independently, and what happens when multi-computations conflict. A further distinction of all of them from cognitivism in linguistics, for example,
\cite{Crof:01,Fauc:02,Lang:08,GoldbergA:13a}, is more agnostic treatment of computation by the latter.} which in the case of CCG implies that semantic representation in the form of logical command relations is insufficient but necessary to acquire the meaning distinctions above.\footnote{Recent theorizing 
in Chomskyan way of thinking tries to unify language of thought with language  (although Chomsky himself has been equivocal in this respect), by identifying gaps of ungrammatical but acceptable expressions on one side, and
ungrammatical and unacceptable expressions on the other, suggesting
that the difference evinces for (I)nternal-language as LOT; see \cite{dupr:20}. The I-language$\approx$LOT hypothesis would face the same
burden of explaining how the surface-syntactically identically behaving expressions can differ in meaning. Both views would have to assume
something unlearned to bootstrap what is learned, to explain the limited and uniform time window of acquisition without infinite regress, which is perhaps
the main spirit of \cite{Fodo:75}, and, in the case of CCG,
what is learned from has no latent computational structure.

And let's not forget that a deeply rooted anthropocentrism about thinking would follow from the I-language$\approx$LOT project. If thought is language, not merely language-like, then animals including closest cousins of ours which lack linguistic ability would imply that they lack thinking as well.
It seems like good minds such as Hume, who wrote back in
\citeyear{hume:echu} that ``the animal infers some fact beyond what immediately strikes his senses''---and by \emph{infer} he did not mean `reason' neither for animals nor humans, deserves a second chance about sources of thought \citep{fodo:03}.}

In addition to following the tradition of assuming that surface structure (which was a theoretical proposal by early Chomsky) is syntactic structure, which
was anticipated by \cite{huss:00}, CCG further proposes that semantics is insufficient but necessary to explain language acquisition. The rationale is the following: No one has been able 
to show convergence to a competence grammar without  an assumption of a homomorphic conceptual base which is available directly to all children in a limited time window \citep{Chom:65,lenn:67,Pink:79,Berw:85,Yang:06,Ambr:14}.
Semantic bootstrapping as an umbrella term covering this way of thinking
assumes that all human languages can be supported by a conceptual structure if it is sufficiently homomorphic to world's syntaxes. What is acquired is syntax-semantics correspondences of words, and what the child is assumed to be exposed to is utterance meanings and forms, not the  words themselves in isolation, and under uncertainty. In other words, the child already knows that forms have meanings, otherwise, if that property is also learned, then we would have had to have seen much greater variation in timeline of language acquisition.
What the child learns is which forms go with which meanings.
Working with this rationale shows signs
of capturing timeline of child development and known patterns, such as vocabulary spurt, sudden jumps, fast mapping, and syntactic bootstrapping (as an effect rather than cause). In this way of exploration
the task of the modeler is not to commit to a specific shape and style of the homomorphic base (logic, psychosemantics, etc.), but to assume that one exists and supports compositionality of concepts and thoughts \citep{Ambr:14}. 
Therefore the critical position that CCG takes is claiming that syntax is learned from semantics at the level of words eventhough they are not received in isolation; see \cite{Aben:15a}. What is variant and what is invariant in this exploration are therefore critical.

\subsection{Invariants and variants of combination}\label{sec:ccg}
The direct connection to sequencing and its semantics in CCG is that bounded constructions need no more than function application, which also happens to be the only primitive of the idea of combinators, and unbounded constructions need no more than composition (\combb) and substitution (\combs). Altogether they amount to Steedman's (\citeyear{steedman00}) principle of projective dependency for language, that surface structure in all its aspects is projected from a lexicalized grammar unchanged, undeleted and unmoved, including avoidance of predicate abstractions on the fly---therefore all predicates are specified in the lexicon. Languages 
differ only in their lexicons.

We have seen these functionalities of planned action as application\xref{ex:pl-fa}, composition\xref{rul:fc}, and substitution\xref{rul:sub}. Linguistic valuation of their input categories is naturally expected to differ from that of  planning, usually considered to consist minimally of the noun phrase (\cat{NP}), the verb phrase (\cat{VP}) and the clause (\cat{S}). For plan valuation I have used \plancat for category assignment. For linguistic categories I use CCG notation for it, which is :=.

These linguistic categories are not considered universal, since everything that is specified in the lexicon can be projected.\footnote{This is one major  difference from Chomsky's universal grammar, where N(oun) and V(erb) are considered innate, treating them as universals, among others such as adjective \citep{bake:03}.} They are seen more of capturing distributional diversity of categorization across the world's languages, held together by universal projection.\footnote{\label{fn:ccg}See \cite{Bozsahin:12,stee:17}
for what would be at stake linguistically if we include rules such as subcomposition\xref{rul:subc}  for projection and similarly-inspired rules, and for some alternatives. Briefly, \combd\,would be required
to syntactically handle examples such as \emph{the film's ending \lbr that I\/\rbr\, and \lbr that you\rbr\,preferred} (Steedman, p.c.). So far, all the syntactic contexts that have been identified requiring 
\combd\,involves one closed-class element such as `that' in this example, for example `what' in \emph{\lbr what you can\rbr\, and \lbr what you won't\/\rbr\, sell}, therefore alternative analyses in which
these elements lexically compose would be available---they are morphologically bound
elements in many languages. 

The other candidate extension for projection, christened \combl\,by Steedman and Boz\c{s}ahin for language (see \cite{stee:17}), which we used implicitly without a name in\xref{rul:s-like}, concerns the case theory of CCG and does not engender predicate abstractions during projection. It would be at work in deriving the fragments in brackets
in the following example: \emph{I persuaded \lbr Mary to buy\rbr\,and~\lbr Harry to sell\rbr\,the house.}
Its connection to \combs\,of\xref{rul:sub} cannot be overlooked in relation to intensionality. \combl\,is defined in one directional variant as follows, cf.\,\combs\,for the same directionality (we shall soon see that $<$\combl~is needed in analyzing expressions such as above):

\cat{(X\fs Y)\fs Z}\lf{f}~~\cat{Y\fs W}\lf{g}~~$\Rightarrow$~~\cat{(X\fs Z)\fs W}\lf{\lambda x\lambda y.fy(gx)}\hfill($>$\combl)

\cat{(X\fs Y)\fs Z}\lf{f}~~\cat{Y\fs Z}\lf{g}~~~$\Rightarrow$~~\cat{X\fs Z}\lf{\lambda x.fx(gx)}\hfill($>$\combs)

The common pattern is that \combs\,and \combl's main functors are necessarily two-argument functions, unlike \combb. One of these arguments is more liberal in \combl, allowing limited intercalation, such as \cat{(A\fs B)\fs C~B\fs D~D~C}, but not \cat{(A\fs B)\fs C~B~C}, adding more to the repertoire of semantically connectable sequences. This is different than the freely order-inverting \combc, which does the intercalation that we just rejected, which is a function of type $\lambda f\lambda g\lambda a\Rightarrow fag$. It is known to overgenerate in language; see \cite{Bozsahin:12} for extensive discussion.} 

Going back to\xref{eg:intro}, repeated below, we can see complex categorization at work in\xref{cat:intro} in controlling semantic nondecomposability
and syntactic decomposability. 
\pex[exno=\ref{eg:intro}]
 \a\begingl
 \gla Zh\=angs\=an  \underline{sh\=eng} \underline{q\`{\i}} le//
 \glb Zhangsan generate air \textsc{asp}//
 \glft  `Zhangsan got angry.'//
\endgl
 \a\begingl
 \gla Zh\=angs\=an  \underline{sh\=eng}  le  h\v{a}od\`{a}  de \underline{q\`{\i}}//
 \glb Zhangsan generate  \textsc{asp}  huge \textsc{nom}  air//
 \glft `Zhangsan got very angry.' (lit. `Zhangsan generated huge air.')//
 \endgl
 \a\begingl
 \gla Zh\=angs\=an \underline{sh\=eng}  w\'an \underline{q\`{\i}} le//
 \glb Zhangsan generate  finish  air  \textsc{asp}//
 \glft `Zhangsan stop being angry.' (lit. `Zhangsan finished generating air.')//
 \endgl
\xe

\pex\label{cat:intro}
\a sh\=eng := \cat{(S\bs NP)\fs\cgs{NP}{\mbox{\scriptsize q\`{\i}}}} \lf{\lambda x\lambda y.\so{angry}_{\circ x} y}
\a sh\=eng := \cat{(S\bs NP)\fs NP}\lf{\lambda x\lambda y.\so{generate}xy}
\xe
where `{\small $\circ$}' as the subscript of a predicate means that what follows it is an event modality of the predicate rather than an argument of the predicate itself, from \cite{bozs:22-jlli}. The correspondence is kept strictly compositional by avoiding vacuous abstractions for example on `\fs \cgs{NP}{\mbox{\scriptsize q\`{\i}}}', also avoiding the
need for implicature when it is not necessary.\footnote{In
\cite{bozs:22-jlli}, a category such as \cgs{NP}{\mbox{\scriptsize q\`{\i}}} is called a head-dependent type, which differs
from an ordinary \cat{NP} type by being all about reference to the head-form \emph{q\`{\i}}. Such types, and even
more specific types such as singletons---references to a token rather than token type as in `bucket' of kick the bucket, need
no auxiliary projection apparatus \citep{bozs:22-jlli}.}

Although decomposable meanings of `generate' and `air' are available in freer syntax of\xreff{eg:intro}{b--c}, the reference of the events
is to becoming angry in all cases, which is not obtained as implicature, that is, by post interpretation.
It is the verbal subcategorization which fixes this reference; cf. freer reference of the verb for `generate' in (b) where cooccurrence with phrases involving
\emph{q\`{\i}} is not expected.
In such categorization and subcategorization we can avoid the polysemy fallacy pointed out earlier (footnote~\ref{fn:fallacy}) by principally deciding
categorization based on categorial choice of reference, as done by main predicate constants in\xref{cat:intro}.
Needless to say, lexical semantics has to play a role in it as well.\footnote{One more process is needed to make categories such as\xreff{cat:intro}{a--b} contrastive, that is, relatively exclusive, to do justice to lexical semantics work that is needed to carve them out. Default reasoning
\citep{Reit:01} is designed to handle inferences of this kind.}

I am not suggesting that CCG's current way of theorizing and complex categorization are complete or minimal in their apparatus. 
I am arguing that a mathematical understanding of the sources of empirical assumptions goes a long
way in appreciating and assessing any linguistic theory. Linguistics is not mathematics, but mathematical understanding
of linguistic structure may reveal the structure of the invariant so that the variants (the linguistic data and its typology) can be better understood.

For CCG in particular, we can have a better understanding
of what is at stake regarding choice of combinational universals (i.e. its necessary apparatus to  deal with all constructions) and projection (sufficiently minimal apparatus for them), by having a closer
look at \combl\,and \combs, respectively.

Combinator \combl\,of footnote~\ref{fn:ccg} has two applications deferred, one to
receive \cat{\fs Z}, and the other to receive \cat{\fs W}. Its combined result carries over these types without reduction.
Its result combines with \cat{W} first, which does not necessarily follow from the deferred applications. In that sense, its combination would  not be necessarily functional; it could be conceived
as combining first with \cat{Z}. 

Moreover, syntactic case functions of some  residuated complement clauses would remain  open problems without the harmonic backward \combl, which is
the backward variant of \combl:
\ex
\cat{Y\bs W}\lf{g}~\cat{(X\bs Y)\fs Z}\lf{f}$\Rightarrow$\cat{(X\fs Z)\bs W}\lf{\lambda x\lambda y.fy(gx)}\hfill($<$\combl) 
\xe 

It is harmonic because the argument which is composed over (\cat{W}) and 
the function composed over (\cat{Y})  have the same directionality.
Two examples from verb-medial languages English and Mandarin Chinese are (only \emph{persuade}'s category is shown for brevity):
\pex\label{ex:l-contexts}
\a I persuaded$_{\cat{((S\bs NP)\fs VP)\fs NP}}$ \lbr Mary to buy\rbr~and \lbr Harry to sell\rbr~the book
\a\begingl
\gla N\`a-b\v{e}n  w\v{o}  h\`ui  sh\`uif\'u 
{\lbr Zh\=angs\=an}  q\`u  {m\v{a}i\rbr} \'erqi\v{e}  L\v{\i}s\`i  q\`u  {m\`ai\,}
  de    sh\=u//
\glb That-Measure  I    will  persuade  Z   to    buy   and     L    to   sell  {rel}  book//
\glft `the books which I will persuade Zhangsan to buy and Lisi to sell'//
\endgl
\xe
For example, (a) in the following analysis shows that \combl\,must be at work
in assembling \emph{Mary to buy} of\xreff{ex:l-contexts}{a} locally so that all verbs and
residual complements can take their arguments with correct cases (of who does what to whom in \emph{persuade} and \emph{buy}, as shown in (b)):\footnote{I follow the CCG convention
of displaying a tree with leaves on top and root at the bottom. A line is drawn to show
span of local combination, for example \emph{Mary} combines with \emph{to buy} in\xreff{der:l-contexts}{a}
using backward harmonic  variant of \combl.  Tree display is flattened to save space.}
\pex\label{der:l-contexts}
\a{\small\cgex{3}{
\lbr Mary & to &  buy \rbr\\
\cgul & \cgul & \cgul\\
\cat{((S\bs NP)\fs VP)\bs(((S\bs NP)\fs VP)\fs NP)} &  \cat{((S\bs NP)\bs((S\bs NP)\fs VP))\fs VP}  & \cat{VP\fs NP}\\ 
\lf{\lambda p.p\,\so{mary}}
& \lf{\lambda p\lambda q.qp}
& \lf{\lambda x_1\lambda x_2.\so{buy}x_1x_2} \\
&\cgline{2}{\cgfc}\\ & \cgres{2}{\cat{((S\bs NP)\bs((S\bs NP)\fs VP))\fs NP}}\\
& \cgres{2}{\lf{\lambda z\lambda q.q(\lambda x_2.\so{buy}\,z\,x_2)}}\\
\cgline{3}{$<$\combl}\\
\cgres{3}{((S\bs NP)\fs NP)\bs(((S\bs NP)\fs VP)\fs NP)}\\
\cgres{3}{\lf{\lambda p\lambda x\lambda x_2.p\,\so{mary}(\so{buy}x\,x_2)}}
 }}\medskip
 
 \a{\small\cgex{2}{persuaded & Mary to buy\\
 \cgul & \cgline{1}{$<$\combl}\\
 \cat{((S\bs NP)\fs VP)\fs NP} & \cat{((S\bs NP)\fs NP)\bs(((S\bs NP)\fs VP)\fs NP)} \\
 \lf{\lambda x\lambda p\lambda y.\so{persuade}(px)xy} 
 & \lf{\lambda p\lambda x\lambda x_2.p\,\so{mary}(\so{buy}x\,x_2)} \\
 \cgline{2}{\cgba}\\
 \cgres{2}{\cat{(S\bs NP)\fs NP}}\\
 \cgres{2}{\lf{\lambda x\lambda y.\so{persuade}(\so{buy}x\,\so{mary})\so{mary}\,y}}
 }}
 \xe

No other CCG projective combinator can combine (a) while maintaining case functions true to their argument-takers, the verbs.

The appeal of \combl\,is that it is known that constructions interact to interleave
their argument taking. If we assume following \cite{steedman96} that some English PPs must  be arguments
\emph{because} they can cause a verb to suspend coordination and complementation
much like node raising, as in (a), then analyses such as (b)
are naturally offered by \combl\,for any such interaction.
\pex\label{ex:carpet}
\a I folded the rug over, and the curtains under, the painting.
\hfill\citealt{steedman96}:41\\
\a \hspace*{-1em}\cgex{3}{folded & the rug & over \\
\cgul & \cgul & \cgul\\
\cgf{(S\bs NP)\fs PP\fs NP} & \cgf{((S\bs NP)\fs PP)\bs((S\bs NP)\fs PP\fs NP)} & \cgf{((S\bs N
P)\bs((S\bs NP)\fs PP))\fs NP}\\
\lf{\lambda x\lambda y\lambda z.\so{fold}yxz} &
\lf{\lambda p.p\,\so{rug}} &
\lf{\lambda p\lambda q.q (\so{over}p)}\\
& \cgline{2}{$<$\combl}\\
& \cgres{2}{\cat{((S\bs NP)\fs NP)\bs((S\bs NP)\fs PP\fs NP)}}\\
& \cgres{2}{\lf{\lambda f\lambda g.f\,\so{rug}(\so{over}g)}}\\
}
\xe

The phenomenon is crosslinguistically and cross-constructionally available. The following Turkish ``split genitive'' (interleaving of \emph{man's house} and \emph{me}) is not entirely intonation-neutral
in my Turkish, therefore we can think of an analysis without \combl; but, for those
who see no difference from the unsplit genitive, \combl\,could provide an analysis:\footnote{This distinction is important. If \combl\,is a freely-projecting combinator, it cannot add such semantics.}
 \ex
\begin{ccgg}{4}{Adam-\i n & ben-i & ev-i & etkiledi.}
{man{-gen.3sg} & I{-acc} & house{-poss.3s} & impressed}
{
\cat{NP\fs NP} & \cgs{NP}{acc} 
& \cat{(S\fs(S\bs NP))\bs(NP\fs NP)} & \cat{(S\bs NP)\bs\cgs{NP}{acc}}\\
\lf{\lambda x.\so{poss}\,x\,\so{man}} & \lf{me} & 
\lf{\lambda q\lambda p.p(q\,\so{house})}
& \lf{\lambda x\lambda y.\so{impress}xy}\\
&&\cgline{2}{$>$\combl$_{\times}$}\\
&&\cgres{2}{\cat{S\bs(NP\fs NP)\bs\cgs{NP}{acc}}}\\
&&\cgres{2}{\lf{\lambda x\lambda y.\so{impress}x(y\,\so{house})}}\\
& \cgline{3}{\cgba}\\ & \cgres{3}{\cat{S\bs(NP\fs NP)}}\\
&\cgres{3}{\lf{\lambda y.\so{impress}\so{me}(y\,\so{house})}}\\
\cgline{4}{\cgba}\\ \cgres{4}{\cat{S}\lf{\so{impress}\so{me}(\so{poss}\,\so{house}\so{man})}}\\
\mc{4}{`The man's house impressed me.'}\\
\mc{4}{`It was the man's house that impressed me.'}
}
\end{ccgg}
\xe

The split is not construction-specific in Turkish. In the following example, \emph{we surrounded} intercalates \emph{Hasan believe}:
\ex
{\small
\cgex{6}{Hasan'\i n& biz-i & ku\c{s}att\i(-k)& san & -d\i\u{g}\i
& \hspace*{-1em}\"universite \\
H{-3s} &  we{-acc} & surrounded{-1p} & believe & {rel.3s} & univ.\\
\cgul & \cgul & \cgul & \cgul & \cgul\\
\cgf{\cgs{S}{3s}\fs\cgs{IV}{3s}}
& \cgf{IV\fs(IV\fs\cgs{NP}{acc})}
& \cgf{VP\bs\cgs{NP}{acc}}
& \cgf{(IV\bs VP)\fs\cgs{NP}{acc}}
& \cgf{(NP\fs NP)\bs(\cgs{S}{3s}\us\cgs{NP}{})}\\
&&\cgline{2}{$<$\combl} \\ &&\cgres{2}{\cat{(IV\fs\cgs{NP}{acc})\bs\cgs{NP}{acc}}}
\\
& \cgline{3}{\cgfx} \\ & \cgres{3}{\cat{IV\bs\cgs{NP}{acc}}}\\
\cgline{4}{\cgfx} \\  \cgres{4}{\cat{\cgs{S}{3s}\bs\cgs{NP}{acc}}}\\
\cgline{5}{\cgba}\\ \cgres{5}{\cat{NP\fs NP}}\\
\mc{6}{`the university which Hasan believes we surrounded'}}}
\xe
Unlike the split genitive, there is no grammatical variant where \emph{bizi} would not intercalate:
*\emph{Bizi Hasan'\i n ku\c{s}att\i(-k) san-d\i\u{g}\i\ \"universite}.
This has been an open problem in Turkish linguistics; see for example \cite{kornfilt88,korn:07,moor:98}.

To recapitulate the discussion of \combl\,for CCG in particular, and for completeness of mathematics of intercalation in general in the context of surface configurationality, it is clear that
disharmonic varieties of \combl\,are nonprojective of command relations \citep{stee:20}. 

\combl~as a combinator would be  doubtful in the analytic base of natural language because
not all variants of \combl\,are on the same ground with respect to Baxter properties. 
This is in sharp contrast with \combb\,and \combs\,because both harmonic and disharmonic varieties of \combb\,and \combs\,satisfy Baxter's condition on universal hierarchies,\footnote{To see this informally, let us assume that in the terminology of \S\ref{sec:baxter}, any result that can locally yield a tree would be considered locally meaningful. We will use sequence description of \S\ref{sec:baxter} in lieu of categories for this; for example, position 1 will yield the overall result of the entire sequence, that is, it must
be some function \cat{1\us A} where \cat{A} is likewise  positions in the sequence turned into categories and `\,\us\,' is some directionality. Because \combl\,abstracts over two arguments we need a wider context to study the impossible Baxter hierarchies  2413 and 3142. 

Take 24135 and 53142. 1's in the these sequences will
be typed as giving the overall result for the tree. All others must appear as argument categories
consistent with the directionality in the sequence to give rise to a local tree for the entire sequence, following the enumeration of \cite{stan:20} for
linguistic examples. The following combinations
show that using disharmonic \combl\,would produce analyses of illicit 24135 and 53142.

\cgex{5}{~~~2~~~ & ~~~4~~~ & ~~~1~~~ & ~~~3~~~& ~~~5~~~\\
\cglines{5}\\
\cat{2} & \cat{4\fs 5} & \cat{(1\bs 4)\fs 3} & \cat{3\bs 2} & \cat{5}\\
&&\cgline{2}{\cgfx} \\ &&\cgres{2}{\cat{(1\bs 4)\bs 2}}\\
& \cgline{3}{$<$\combl$_\times$}\\
&  \cgres{3}{\cat{(1\bs 2)\fs 5}}\\
& \cgline{4}{\cgfa} \\ & \cgres{4}{\cat{1\bs 2}}\\
\cgline{5}{\cgba}\\ \cgres{5}{1}
}\hspace*{4em}
\cgex{5}{~~~5~~~ & ~~~3~~~ & ~~~1~~~ & ~~~4~~~& ~~~2~~~\\
\cglines{5}\\
\cat{5} & \cat{3\fs 2} & \cat{(1\fs 4)\bs 3}  &  \cat{4\bs 5}  &\cat{2} \\
&\cgline{2}{\cgbx} \\ &\cgres{2}{\cat{(1\fs 4)\fs 2}}\\
& \cgline{3}{$>$\combl$_\times$}\\
&  \cgres{3}{\cat{(1\fs 2)\bs 5}}\\
\cgline{4}{\cgba} \\ \cgres{4}{\cat{1\fs 2}}\\
\cgline{5}{\cgfa}\\ \cgres{5}{1}
}
}
 which leaves all the preceding examples in want of analyses.
 
 Turning to combinator \combs\,and to the question of minimality of projection, we can see that its dependency projection is more than what is basically required, which is handling one dependency at a time in merging constituents; recall\xreff{ex:gaps}{i}.
 
 Empirically, the syntactic world that is crosslinguistically carved by \combs\,shows all the signs of a headed construction, that is, as something that does
 not need free projection, in other words, an invariant.
 
CCG's \combs\, analyzes parasitic gaps \citep{Ross:67,Bres:77,Tara:79,Chom:82,Engd:83}, for example
\emph{the articles that I filed \_ without reading \_$_p$}, where `\_$_p$' is the parasitic gap. It is considered to be parasitic on the first gap because, without the first gap,
we would get ungrammaticality:
*\emph{the articles that I filed them without reading}. 
Its categorial syntax-semantics was discovered by \cite{Szab:83} for the special case of the category of VP, and identified by \cite{Stee:87} to be an instance of \combs, in this particular case for the backward crossing variety.

It is a projecting combinator of CCG that handles multiple dependencies in one step. We need a closer look to see this property:

\ex\label{ex:pg-ccg}
\cgex{3}{
filed & without & reading\\
\cglines{3}\\
\cat{(S\bs NP)\fs NP}
& \cat{((S\bs NP)\bs(S\bs NP))\fs\cgs{VP}{ing}}
& \cat{\cgs{VP}{ing}\fs NP}\\
 \lf{\lambda x\lambda y.\so{file}xy}
 & ~~~\lf{\lambda p\lambda q\lambda x.\so{wout}(px)(qx)}
 & ~~~\lf{\lambda x\lambda y.\so{read}xy}\\
& \cgline{2}{\cgfc}\\
& \cgres{2}{\cat{((S\bs NP)\bs(S\bs NP))\fs NP} \lf{\lambda p\lambda q\lambda x.\so{wout}(\so{read}px)(qx)}}\\
\cgline{3}{\cgsbx}\\
\cgres{3}{\cat{(S\bs NP)\fs NP} \lf{\lambda w\lambda x.\so{wout}(\so{read}wx)(\so{file}wx))}}
}
\xe

Notice that it is not the lexically specified $x$ dependency emanating from \emph{without}
which \combs~handles; it is the one within $p$ and $q$ of \emph{without} symbolized by $w$ in the bottom row. No other CCG combinator doubly peeks deep in one step.

This is altogether different than lexically specified multiple dependencies, for example control verbs. If we think of an object-control verb
 as having the following category, the $x$'s in the predicate-argument structure may look like similarly multiple use; however, this is controlled by an  element of grammar, namely \emph{persuade}, much like $x$ of \emph{without} above. 
\ex
persuade := \cat{(S\bs NP)\fs VP\fs NP}\lf{\lambda x\lambda p\lambda y.\so{persuade}(px)xy}, 
\xe

In the coordination of multiple clauses with multiple dependencies, the situation is same as that of \emph{persuade}: the head
of conjunction, not multiple projection of multiple dependencies, handles the correspondence, for example for English with the category below when \cat{X}s are residual ($p$ and $q$ being $\lambda x.\cdots x$):\footnote{We are here assuming pointwise recursion on the semantic
types of \cat{X} at the level of the ground element, following \cite{Gazd:80,Part:83,Stee:11a}.}
\ex
{and} := \cat{(X\bs X)\fs X}\lf{\lambda p\lambda q.\so{and}(p)(q)}
\xe

We can see the difference made by lexical elements even in double-control constructions:
\ex\label{ex:arka}
\begingl
\gla {\rightcomment{Indonesian}Mobil} mana yang kau=coba {\lbr~jual~\rbr}//
\glb car which {FOC} {2sg=Uv}.try {UV}.sell//
\glft `Which car did you try to sell?' \trailingcitation{\citealt{arka:14}:(2)}//
\endgl
\xe

Assuming we have \emph{coba} `try' := \cat{(S\bs NP)\bs\cgs{NP}{pro}\fs(VP\fs NP)}\lf{\lambda p\lambda x\lambda y.\so{try}(pyx)x},
with conditions on embedded subject being matrix pronominal due to preposing \citep{arka:98}, multiple-control is lexically mediated by the control verb,
therefore it is not a case of handling multiple dependencies in one analytic step.
(It is considered multiple-control because the syntactic correspondents
of both $x$ and $y$, `\cat{\bs\cgs{NP}{pro}}' and `\cat{\bs NP}', are realized in the matrix clause.)\footnote{We are following \cite{kana:86} in assuming
that Indonesian is basically SVO. In the AV voice \emph{coba} is realized as \emph{mencoba}, and it is ordinary
\cat{VP} control:
\keepexcntlocal
\ex[exno=i,numoffset=2em]
\begingl
\gla Aku {mencoba$_{\cat{(S\bs NP)\fs VP}}$} {\lbr menjual} mobil {itu\rbr}.//
\glb {1sg} {AV}.try  {AV}.sell car that//
\glft `I tried to sell the car.' \trailingcitation{\citealt{arka:14}:(9a)}//
\endgl
\xe

In double control, both matrix and complements verbs are UV-voiced, where `car' 
in\xref{ex:arka} is syntactic subject in the matrix clause and in the complement clause according to
\citeauthor{arka:14}. My analysis of double control is based on his analysis: `car which {FOC}' would be type-raised as syntactic subject in an SVO language, that is, as
\cat{S\fs(S\bs NP)}. I do not analyze double control as object-extraction, concurring with Arka. This subject corresponds---without an additional layer of argument structure---to the placeholder
for the commanded argument of `sell'; it is not the actor. In this sense I believe the pronominal restriction on shared agent argument `you'
points to non-actor preposing.}

In summary, lexical double-use of a resource is not what \combs\,does. \combs\,is  multiple dependency
handling in one step of projection. We must find languages where this is necessary, not just sufficient.

There is plethora of analyses for parasitic gaps, and
further assessment of multiple dependency projection is needed. \cite{culi:01} is a comprehensive volume on the subject. \combs-analysis of the construction implies that it is not a species of coordination. \cite{will:90} however notes that the parasitic gap of the parasitic gap construction is always optional (e.g. \emph{the articles that I filed without reading them}), therefore \emph{before} in\xref{ex:willpg} can be treated as coordinating or subordinating conjunction licensing both examples (also, \emph{without}):
\pex\label{ex:willpg}
\a Who would you warn \_ before striking \_ \trailingcitation{\citealt{will:90}:265}
\a Who would you warn \_ before striking him
\xe

Optionality is true in cases where offline judgment tests and online reading have been used to justify
 multiple dependency handling in one fell swoop:
\ex
What did the attempt to repair \_ ultimately damage \_/it ? \hfill adapted from \citealt{Phillips:06}:(3c)
\xe

Williams also notes that there is a wide range of acceptability in English parasitic gap environments.
Therefore an analytic \combs~implying universal access to it seems questionable. 

I take Williams's assessment that, as one goes down the list below, from most acceptable to least, the structure is less and less coordinate-like, to be
an indicator that it must be a headed construction, not necessarily by true coordination (see \cite{niin:10} for some reasons for that), but in any case not part of invariant projection. 
\pex
\a Who did you meet and dislike \trailingcitation{\citealt{will:90}:279}
\a What did you file before reading
\a The man who people who meet like
\a Who would pictures of upset
\a Who did you promise friends of to try to find
\xe 

For example, \cite{kath:01} notes that the parasitic gap construction in German is 
lexically headed by a select set of prepositions.\footnote{A sublexical head for the construction is also a possibility which has been realized.
In Turkish, I cannot think
of a parasitic gap configuration without the involvement of the ablative \emph{-den} `from' suffix, with or without verbal negation. It has been noted earlier, for example by \cite{Engd:83}, that languages like Turkish can opt out of the construction because of optional pro-drop. It seems that we must concentrate on consequences of the headedness of the construction by a closed class of grammatical elements such as affixes and function words.}

In summary, there seems to be good reasons to consider an intercalating lingustic invariant 
to be not empirically adequate, and simultaneous multiple dependency projection to be construction-specific, that is, variant, across languages.

\subsection{Mobile logical form and grammar}
With a mathematical understanding of the proposed invariants and why some of them can be considered variant, we can have a closer look at the most important variant for linguistic action, the verb. Verbal categories materialize as follows using mobile semantics of \S\ref{sec:mobile}, for example for \emph{persuade}:
\pex\label{ex:persuade}
\a persuaded := \cat{((S\bs NP)\fs VP)\fs NP}\lf{\lambda x\lambda p\lambda y.\so{persuade}.\pize\,(px)xy}
\a \cgex{1}{persuaded\\
\cgul\\
\cat{((S\bs NP)\fs VP)\fs NP}\lf{\lambda x\lambda p\lambda y.\so{persuade}.\pize\,(px)xy}
}
\xe
In (a) we use the declarative CCG notation, where to the left of `:=' is the surface form of the verb. We have been displaying such categories like (b) in analysis. 

The mobile asynchrony, or offlineness, is invariably part of all lexical mobile logical forms as outcome of their semantics, using `\pize' as a suffix in all linguistic predication. (Note that $x,p,y$ are placeholders, respectively corresponding to \cat{\fs NP}, \cat{\fs VP}, and \cat{\bs NP} because
of the order of specification, not semantic constants of \emph{persuade}.  As such they cannot receive `\pize'.)

With this categorization, the bounded constructions in which this verb can participate, and the unbounded ones, unfold as follows in the course of the surface-analytic process of single computation:
\pex\label{der:persuade}
\a{\small\begin{ccg}{5}{Mary & persuaded & John &  to &study}
{
\cat{Mary} & 
\cat{((S\bs NP)\fs VP)\fs NP} & \cat{NP}  & 
\cat{VP\fs IV}  & 
\cat{IV}\\
\lf{\so{mary}.\pize} &
\lf{\lambda x\lambda p\lambda y.\so{persuade}.\pize\,(px)xy} &
\lf{\so{john}.\pize} &
\lf{\lambda p.p} &
\lf{\lambda x.\so{study}.\pize\,x}\\
 & \cgline{2}{\cgfa}\\ &\cgres{2}{\cat{(S\bs NP)\fs VP}\lf{\lambda p\lambda y.\so{persuade}.\pize\,(p\,\so{john}.\pize)\so{john}.\pize\,y}}\\
 &&&\cgline{2}{\cgfa}\\ &&&\cgres{2}{\cat{VP}\lf{\lambda x.\so{study}.\pize\,x}}\\
 & \cgline{4}{\cgfa}\\ &\cgres{4}{\cat{S\bs NP}\lf{\lambda y.\so{persuade}.\pize\,(\so{study}.\pize\,\so{john}.\pize)\so{john}.\pize\,y}}\\
 \cgline{5}{\cgba}\\ \cgres{5}{\cat{S}\lf{\so{persuade}.\pize\,(\so{study}.\pize\,\,\so{john}.\pize)\so{john}.\pize\,\so{mary}.\pize}}
}
\end{ccg}}

\a \hspace*{-1.5em}{\scriptsize\begin{ccg}{7}{This man,& Jenny & knows & Mary & persuaded & to &study}
{
\cat{S\fs(S\fs\cgs{NP}{+topic})} &
\cat{NP} &
\cat{(S\bs NP)\fs S} &
\cat{NP}  & 
\cat{((S\bs NP)\fs VP)\fs NP} & 
\cat{VP\fs IV}  & 
\cat{IV}\\
\lf{\lambda p.p} &
\lf{\so{jenny}.\pize} &
\lf{\lambda p\lambda x.} &
\lf{\so{mary}.\pize}  &
\lf{\lambda x\lambda p\lambda y.} &
\lf{\lambda p.p} &
\lf{\lambda x.}\\
\blf{(\so{def}.\pize\,\,\so{man}.\pize}&&\blf{\so{know}.\pize\,p\,x} &&
\blf{\so{persuade}.\pize\,(px)xy} & &
\blf{\so{study}.\pize\,x}\\
\bm{\wedge\,\so{topic}.\pize\,\,\so{man}.\pize)}&&&&&\cgline{2}{\cgfa}\\
&&&&&\cgres{2}{\cat{VP}\lf{\lambda x.\so{study}.\pize\,x}}\\
&&&&&\cgline{2}{\cgbtr}\\
&&&&&\cgres{2}{\cat{(S\bs NP)\bs((S\bs NP)\fs VP)}}\\
&&&&&\cgres{2}{\lf{\lambda p.p(\lambda x.\so{study}.\pize\,x})}\\
&&&&\cgline{3}{\cgbx}\\ 
&&&&\cgres{3}{\cat{(S\bs NP)\fs NP}}\\
&&&&\cgres{3}{\lf{\lambda x\lambda y.\so{persuade}.\pize(\so{study}.\pize\,x)xy}}\\
&&&\cgline{1}{\cgftr}\\
&&&\cgres{1}{\cat{S\fs(S\bs NP)}}\\
&&&\cgres{1}{\lf{\lambda p.p\,\so{mary}.\pize}}\\
&&&\cgline{4}{\cgfc}\\ &&&\cgres{4}{\cat{S\fs NP}}\\
&&\cgres{5}{\lf{\lambda x.\so{persuade}.\pize(\so{study}.\pize\,x)x\,\so{mary}.\pize}}\\
&&\cgline{5}{\cgfc}\\ &&\cgres{5}{\cat{(S\bs NP)\fs NP}}\\
&&\cgres{5}{\lf{\lambda x\lambda y.\so{know}.\pize(\so{persuade}.\pize(\so{study}.\pize\,x)x\,\so{mary}.\pize})\so{y}}\\
&\cgline{1}{\cgftr}\\
&\cgres{1}{\cat{S\fs(S\bs NP)}}\\
&\cgres{1}{\lf{\lambda p.p\,\so{jenny}.\pize}}\\
& \cgline{6}{\cgfc}\\
& \cgres{6}{\cat{S\fs NP}}\\
& \cgres{6}{\lf{\lambda x.\so{know}.\pize(\so{persuade}.\pize(\so{study}.\pize\,x)x\,\so{mary}.\pize)\so{jenny}.\pize}}\\
\cgline{7}{\cgfa}\\
\cgres{7}{\cat{S}\lf{\so{know}.\pize}}\\
\cgres{7}{\bm{(\so{persuade}.\pize(\so{study}.\pize\,(\so{def}.\pize\,\so{man}.\pize\,\wedge\,\so{topic}.\pize\,\so{man}.\pize))(\so{def}.\pize\,\so{man}.\pize\,\wedge\,\so{topic}.\pize\,\so{man}.\pize)\,\,\so{mary}.\pize)\so{jenny}.\pize}}
}
\end{ccg}}
\xe
In (a), it is explicit that John is persuaded and he studies. In (b), this is upheld, for `this man'; 
see\xref{ex:pickup-tree} for the logical command relation and its notation.

All the semantic differences and surface identity of `promise/expect/want' from `persuade' in similar experiences as above
then follows from these verbs having the same surface recipe (syntactic category) but different semantics, which
we write below without the syntactic category to avoid repetition:\footnote{As with planning, the directional variants of rules of combination make a difference in how elements are interpreted. \cite{steedman00} shows that the options for language are principally restricted, to harmonic and
adjacent compositions such as (\cgfc) and (\cgbc), and crossing and adjacent ones such as (\cgfx) and (\cgbx), as in\xref{der:persuade}.}
\ex
\begin{tabular}[t]{lll}
persuaded &:= &\blf{\lambda x\lambda p\lambda y.\so{persuade}.\pize\,(px)xy}\\
promised &:= &\blf{\lambda x\lambda p\lambda y.\so{promise}.\pize\,(py)xy}\\
expected & := &\blf{\lambda x\lambda p\lambda y.\so{expect}.\pize\,(px)y}\\
wanted &:= &\blf{\lambda x\lambda p\lambda y.\so{want}.\pize\,(px)y}
\end{tabular}
\xe

It will be noticed that in\xreff{der:persuade}{b} we make crucial use of another functionality we have seen before: type-raising\xref{t-raise}. In the domain of organized action it corresponds to affordances \citep{Stee:02}.
The CCG claim is that in language it corresponds to all arguments of verbs having case, for example nominative and accusative in Latin,
ergative and absolutive in Dyirbal, and mixed or partial in others depending on verbal categories. Its syntactic category is language dependent, and its semantics is invariant, as shown in\xref{t-raise}.

This means that we could write\xreff{der:persuade}{a} with type-raising too, as below, with same surface and logical command relations, where all arguments are ``cased'' by being type-raised. Now all arguments take their subcategorizers (the verbs) as argument.
\ex
{\scriptsize\begin{ccg}{5}{Mary & persuaded & John &  to &study}
{
\cat{Mary} & 
\cat{((S\bs NP)\fs VP)\fs NP} & \cat{NP}  & 
\cat{VP\fs IV}  & 
\cat{IV}\\
\lf{\so{mary}.\pize} &
\lf{\lambda x\lambda p\lambda y.\so{persuade}.\pize\,(px)xy} &
\lf{\so{john}.\pize} &
\lf{\lambda p.p} &
\lf{\lambda x.\so{study}.\pize\,x}\\
& & \cgline{1}{\cgbtr}\\ && \cgres{1}{((S\bs NP)\fs VP)\bs((S\bs NP)\fs VP)\fs NP)}\\
&&\cgres{1}{\lf{\lambda p.p\,\,\so{john}.\pize}}\\
 & \cgline{2}{\cgba}\\ &\cgres{2}{\cat{(S\bs NP)\fs VP}\lf{\lambda p\lambda y.\so{persuade}.\pize\,(p\,\so{john}.\pize)\so{john}.\pize\,y}}\\
 &&&\cgline{2}{\cgfa}\\ &&&\cgres{2}{\cat{VP}\lf{\lambda x.\so{study}.\pize\,x}}\\
 &&&\cgline{2}{\cgbtr}\\
 &&&\cgres{2}{(S\bs NP)\bs((S\bs NP)\fs VP)}\\
 &&&\cgres{2}{\lf{\lambda p.p(\lambda x.\so{study}.\pize\,x})}\\
 & \cgline{4}{\cgba}\\ &\cgres{4}{\cat{S\bs NP}\lf{\lambda y.\so{persuade}.\pize\,(\so{study}.\pize\,\so{john}.\pize)\so{john}.\pize\,y}}\\
\cgline{1}{\cgftr}\\ \cgres{1}{S\fs(S\bs NP)}\\
\cgres{1}{\lf{\lambda p.p\,\so{mary}.\pize}}\\
 \cgline{5}{\cgfa}\\ \cgres{5}{\cat{S}\lf{\so{persuade}.\pize\,(\so{study}.\pize\,\,\so{john}.\pize)\so{john}.\pize\,\,\so{mary}.\pize}}
}
\end{ccg}}
\xe
It is  important to keep in mind that case  only targets verb-like lexical elements, that is, it is a second-order (i.e. structural) function which only depends on eventful singularities in a lexicon. What makes it an empirical universal is its invariant relation to such singularities.
And it does so for all arguments, nominal and clausal, as above. Notice that the clausal argument \emph{to study} is also cased by being type-raised. 

With this in mind, it must be evident that even in language everything can be treated as a function anyway, including entities and predicates, much like daily organized activity and affordances, therefore offlineness of language only comes from the speakers-hearers deliberately letting go of change in expressing themselves. The asynchrony suffixes that we have added  in this section are reflections of the common understanding that people know the difference between acting and speaking and hearing about acting. Knowing the difference does not preclude them from continuing to track change,
as full representation of for example\xref{ex:persuade} in mobile semantics shows below; cf.\xref{ex:pickup} for organized action.
\ex\label{ex:persuade-mlf}
\begin{tabular}[t]{lll}
persuaded & := & \cat{((S\bs NP)\fs VP)\fs NP} 
\lf{\lambda x\lambda p\lambda y.\so{persuade}.\pize\,(px)xy}\\
&& (\texttt{pre:} $\so{not}\so{do}(x,p), \so{know}(y,x)$\\
&& ~\texttt{add:} $\so{attempt}(x,p), \so{convince}(y,x)$\\
&& ~\texttt{del:} $-$~).\pize
\end{tabular}
\xe
Only the linguistic logical form and semantic conditions of the verbal form are necessarily offlined and consequently asynchronic. Syntax is the active element when computation is driven by the syntactic category, revealing the semantics in locked-step.
We assume that all linguistic elements are of this dual-functional character
and uniquely representational in nature, taking part in a single computation of dealing with change, recording and reporting it. 

Affordances and type-raising have implications for what can be offlined without letting go of tracking change. In the case of langauge,
it is the lexical  verb's projection that must be heeded, otherwise case could not always be a second-order (i.e. structural) function. 

Interesting consequences arise for abstracting over a verb's action or event, that is, making it opaque in content. (Recall that it does not seem to be necessary for language, although generative grammarians might find it useful, employing its functional equivalent: movement.)
It means that \combd\,in action, which we dubbed `subcomposition' in\xref{rul:subc}, can be a spandrel for deceit in onlineness, whereas language would use lexical or surface ambiguity for these functions. This is because it creates predicate abstractions, that is, nonlexical predicates, as a means of covertness in action without needing offlineness of 
language, that is, not having to keep silent about something to be able to do covert action.
We can conceive a common base for language and planning in the simple mathematics of \comba, \combt, and \combb, without sacrificing their differences and potential sources of divergence, for example \combs, \combd\,and~\combl.


\section{The proposed synthesis in perspective}\label{sec:pl}\label{sec:synthesis}
Three mainstream approaches to semantics in linguistics are interpretation of a formal system with model-theoretic truth-conditional semantics
\citep{Mont:70}, interpretation of thematic structure from syntactic structure \citep{Grim:90,Hale:02,bore:05}, and interpretation of surface structure from conceptual structure \citep{talm:88,Jack:90,langacker87}, none of which
addresses the Frame Problem. The problem  is inherently procedural.

We looked at whether there is something to be gained 
if we conceive the main task of semantics as understanding change, both in speaking-signing and acting, which includes tracking it and coping with it.
Tracking requires internal representation of action, online or offline,
and coping requires reference to past, current and future states and ways to deal with them.
The implication is that all knowledge is procedural and prone to side effects. 
For example, understanding a verb means knowing what it changes and what it doesn't, same as understanding an action
means knowing what it changes and what it doesn't. Their categorization is obviously domain-specific, but, their why and how questions appear to target the same problem.

This way of thinking connects language and everyday planning in circumscribed ways to add 
to the debate that  language can first piggyback on then coevolve with some other trait \citep{cool:05,deac:97,toma:99,Jack:05,reul:10,Stee:02,Stee:14e,hrdy:11,Ullm:04}. I adopted a Marrian (\citeyear{marr:77}) strategy and made the  proposal explicit
at the algorithmic (middle) level, to try to avoid pitfalls of cross-level purported dichotomies such as the one in grammar/lexicon, which supposedly corresponds to procedural/declarative distinction at the substrate (lowest) level according to \cite{Ullm:04}. Another level-3 concern is the
 supposed language regions of the substrate or causal role of lateralization in language modularity, which have been 
questioned by \cite{deac:97}. Yet another one is ``desymbolization'' of \cite{reul:10}, 
which means that Chomskyan ``merge'' just works independent of semantics,
which, if taken technically at the level of algorithms, 
cannot support this level if correspondence theories are right. 

My only appeal to Marr's level-1 has been for the purpose of showing that some algorithms
serve to represent offlineness of language  and onlineness of action faithfully. This is concern for the problems rather than their descriptions, without a dichotomy at the algorithmic level. 
For us to assess why expressing thought with language or with action might differ, for a start we must be able to show their differential effects in environment control of agents doing actions versus talking about actions.

To make the proposal explicit for these algorithms, I adopted categorial grammar without multiple dependency projection, which has come to accommodate rich vocabulary of semantic interpretation in a Logical Form (LF), diverging from  exclusively syntactic ways of dealing with LF (see \cite{carl:83} for discussion),   to the point of answering meaningful  questions about an utterance
\citep{LewisM:13a,stee:19ohb}. The LF is the only level of representation in CCG, not for reasons of linguistic-philosophical parsimony but for its empirical
dividends at level-2 and level-3,
by which we can computationally study how children  acquire language \citep{Aben:15a}. In current work
I turn the LF into a Mobile Logical Form (MLF) of \emph{fluents}, i.e. dynamically changing predicates for nonmonotonic reasoning, whose relevance to planning has been known \citep{lifs:02,Stee:02}. I do this by incorporating ideas from mobile communication theory ($\pi$-calculus) of \cite{miln:93}. I make use of it to show that concurrent computation 
when coupled with a locally interpretable grammar, by which I mean a grammar in which all elements are compositional and locally interpretable, does not require multi-level level-3 architectures or separate level-2 computations to incorporate parallelism into language and action, at the same time supporting
language for sequential expression of thought \citep{Fodo:75}.

The central claim is that uniquely human characteristics of language-specific categorization and its acquisition
can be bootstrapped by the ability to go offline, the ability to let go of pursuing change implicated by speech-sign. I argued that  spatiotemporal semantics must move up to mobile semantics to make the connection explicit.
The implication for time course of evolution of the substrate is that mobile meanings and online thinking about plans may be the representational support for offline thinking.
To see the support, how mobile meanings evolve is worth reiterating.

The semantics of $\pi$-calculus that concerned us is the standard reduction rule of concurrent computation, from\xref{def:pisem}:

(\piou{x}{z}$.P$ \pipa\,\,\piin{x}{y}$.Q$) $\Rightarrow$ ($P$ \pipa\,\, $Q\lbr z/y\rbr$)\\
It means that processes $P$ and $Q$, whose timing and duration are not known to each other because `$\!\!\pipa$' means `concurrent',
meet by $P$ supplying $Q$ with an input in the channel $x$. 
($P$ sends and $Q$ receives in this notation.)
What is supplied, $z$, and what is expected, $y$, are local. Semantics of language and planning
consists solely of such processes in their mobile logical forms (MLFs).

Mobile semantics' semantics  is crucial in studying offlineability of language and its connections to planning. 
All MLF objects of language are considered terminal processes (and suffixed as such, with the ``.\pize~process''), whereas no planned action is a terminal process unless deliberately specified as such 
(for example, seeking shelter in dancing in the rain example\xref{ex:dance}),
because the communicator
has to continue to adjust to states of affairs in the world.\footnote{``.\pize'' is a level-2 process, which is part of the algorithmic system. It is related to the level-1 concern for understanding letting go
and not being able to do so.
No claim is made about its physical or psychological grounding at level-3.
One connection is Barsalou's \citeyear{bars:99} ``perceptual simulators'', which are modal rather than amodal top-down \emph{and} bottom-up symbolizations, as befits a level-3 concept. However we  note that the `let go' of change implicated by .\pize~in language does not depend on the success of the simulation.
It successfully terminates, no matter what the world makes of it.
}
I covered some  examples of  MLFs of planned action and language  to support the idea.

Mobility in the logical form brings to discussion the role of semantics in a competence grammar, which is a top level concept according to Marr. Here too I started at the algorithmic (i.e. middle) level
to make it explicit that Putnam's (\citeyear{Putn:75a}) distinction of expert knowledge of meaning of words, say $gold^\prime$ for the word \emph{gold}, and the ``collective linguistic body's'' knowledge of the word, must both be \emph{expressed} 
as ${gold}^\prime.\pize$. This is the socio-linguistic meaning of the word \emph{gold}, and it has nothing to do with  gold expertise, that is, what \so{gold} means to the bearer of that knowledge.
It is social, or \emph{shared} psychological spaces in \citealt{sapi:33}:17-18 terminology (\emph{not} psychological states), in the sense of being necessarily exchanged. It is linguistic in the sense that personal {expressions} are always terminal processes.
The difference is in the procedural semantics of  personal knowledge reflected in  personal meanings of $gold^\prime$ for the common folk and the expert, and that has nothing to do with linguistic competence reflected in \so{gold}.\pize. A representational support to clarify the distinction is necessary but not sufficient.

We might now appear to be at an impasse in taking an internalist or externalist position in relation of semantics to syntax.
The Twin Earth argument \citep{Putn:75a} pointed out that having the same surface expression does not mean having the same thought.
My proposal for studying this phenomenon is to be explicit at the middle level about how not to directly ground linguistic LFs 
in the world (cf. \cite{parte:81,John:83}), and leave it at that, and about how that differs in representation of online action.\footnote{There is a further
dimension to the debate on representationalism in cognitive science, which revolves around the idea of whether representations are empirically necessary
to understand cognition; see for example \cite{Fodo:81,beer:03,miro:12} for the debate. Attacking the problem starting at Marr level-2 
has the implication that positing some representations may leverage bridge theories between levels, and they might allow us to explore 
connections at the problem level rather than solution level. Dynamic systems theorists who tend to eschew positing representations at level-3 can do so mainly because their time series computations at level-3 report their results back to level-2 mechanisms for example in analog computations, which are themselves representational to carry out the computation \citep{bozs:turtles}. \emph{That} mechanics has to be representational.}
A further implication is that, unlike Marconi's (\citeyear{marconi97}) multiple inward/outward competence, called inferential/referential competence, 
we can work with one notion of competence, of connecting syntax and semantics and their projection all the way to surface structure, no more no less. This is analytic competence, much like we argued for for planning in Section~\ref{sec:plans} and for language in Section~\ref{sec:language}. It differs from Chomsky's notion 
of uniqueness of syntactic competence in having to carry terminal processes of semantics in lock-step with syntax. This ability is not acquired from performance, although the knowledge of which meanings go with which forms is.

A grammar and its model can be conceived to be different but causally related species, Vygotsky-style,
even Schopenhauer,
and we can maintain the uniqueness of competence without idealistic interpretations, both for grammars of languages and plans.

\section{Discussion}\label{sec:conclusion}
Conceiving semantics as semantics of change, planning as scripted worldview update in a limited channel
of mobility, and language as terminal semantic processes working under universal syntactic constraints
reveals a problem space, not solution space, for common understanding of language and planning.
This space can therefore be assumed to have already incorporated thoughts, impressions and expressions about the world.

It allows us to ask the following question: Why do we build these views? It must have something
to do with putting all of it to online and offline use, for control of environment.
We can conceive this control as a closed-loop, that is, a system which is always responsive
in its inputs to what it produces, but also an  open-loop, that is, a system 
with some internal preferences for desirable input independent of output, choosing a world to be in rather
than coerced to be in it. In this sense it is genuine control. (e.g. If in my arm I have muscle soreness and I want to limit responses demanded by environment so that it can heal, I can put it
in braces, in effect causing internal restructuring to avoid certain external demands, choosing to be in a certain niche.)
Systems theory and computer science can be combined to address philosophical problems provided we avoid the computer metaphor \citep{cise:99}. I add that we can avoid all metaphors at level-1 and level-2 and study metaphors at level-3, which seems to be a real need. Computer science and control systems theory are there to provide transparent working principles at level-2 to assess claims.

Computer science's appeal in exploring human practice 
is its unique ability to show transparently that a simple mechanism to understand
complex behavior need not itself become complex as the behavior becomes more complex. (Hiding
this transparency, either deliberately or accidentally, is the political force of the computer.)
Transparency is the key in building confidence in a model in this way of thinking. The daredevil spirit of the process (e.g. \emph{procedural epistemology} of \cite{abel:85}) is admittedly difficult to come to grips with for this reason.  However, white-box modeling 
with the computer is the bedrock of computational ideas for understanding complex human behavior, including language and planning, and perhaps even more.

About another daredevil approach in a seemingly unreachable domain, consciousness, Daniel Dennett once said that Julian Jaynes attempted ``software archeology'' in search of origin of consciousness \citep{denn:86}.

\cite{jayn:76} had caused quite a stir in the 1970s, which continues
 these days as a love-or-hate affair with Jaynes. He suggested that human consciousness is a recent event, reported on historical record around second millennium B.C., which piggybacked on language, which piggybacked on unconscious planned action and its admonitoring---in the case of Jaynes by gods, which started as the mind's coping with hallucinations caused by change of habitat and historic upheavals, i.e. loss of god-talk. 

Although we are not concerned with consciousness in the article, and what Jaynes says about its recency is difficult to believe, 
the way he argued is what prompted Dennett to call it \emph{software} archeology. Jaynes appealed to what Dennett calls a ``top-down approach'', meaning building a just-so story and filling in the details transparently about history of natural and cultural developments in trying to make sense of it altogether, from its substrate to problem itself---not just some solution, with a causal link in between. In other words, he practiced what Marr suggested for complex information processing problems, apparently without knowing Marr's work at the time of writing the 1976 book, although by 1978 he knew
about the computational models in similar spirit \citep{jayn:78}, citing \cite{pyly:78} for ``that other method''. 

He made up representations, that is, targeted level-2, to address a level-1 problem, which enabled him to look for its archaeology at level-3. That in turn enabled him to build a historical, genetic and epigenetic account for that level, unlike one biological event causing ``merge'' in humans \citep{bolh:14} to bring about language (and language only).
And something did happen to basal ganglia, parts of which have the major functions 
of cortical synchronization and managing
 ``appropriate sequence of thought'' \citep{brow:98},  6 to 7 million years ago, by which the most distinct genetic difference is in FOXP2 transcriptional factor of chimpanzees on one side, and humans, neanderthals and denisovans on the other. It does not appear to be the unique biological event that \cite{bolh:14} identify with ``merge'', because they do not consider it---rightfully I think---to be ``the gene for language''. It  seems that having a Marr-plus-anthropological account leaves more operating room for understanding Nature doing its work in the course of time.

His explanation of consciousness may be questionable (though not dubious, to \cite{will:10} or to present author, among others). The argument in this article is that his reasoning has been promising for all aspects of cognition. We can conceive a representational basis to computationally support \emph{know-how} of planned action of colloquy of agents, and language can be seen as a further categorial shift where
speech-sign acts are semantically terminal processes at this level, unlike action, which means they are offline. Language can be understood to project only from a grammar and never 
create predicate abstractions on the fly in speech-sign delivery or speech-sign understanding, meaning it only reports what is being offlined, or receives it. In daily action, most processes
cannot be terminal because we have to live with their consequences, i.e. their preconditions, add lists, delete lists, and the state of the world. However, the action may be overt, or it may be
some kind of predicate abstraction, the real intent (and even conditions) of which may be opaque to participants.
This is fertile ground for studying \emph{deceit}, a favorite  subject of Jaynes, and to lay rudiments of a theory for conjuring minds, from animals to human. 

\cite{Fodo:75,Fodo:81,Fodor:08} had hoped to get the best out of computer science in cognition by choosing combinatory explosion as the basis of creativity in language and behavior. Difference in offlineness has not been addressed in this way of thinking as the primary source of complexity, as cause of rise of belief systems. 	This too, I claim, can be deeply computer-friendly to study if we look at computer science as studying extended practice of the human, as suggested also by some Turing Award recipients \citep[e.g.][]{vali:13}, not practice of the extended human. (This is quite different than
other proposals for human in the loop for computers, such as \cite{vard:11}, which can be summed up as ``computing for humans''. The current proposal is ``computing qua humans''---what kind of control and data structures humans can conceive
can be \emph{programmed} by humans to the extent of not needing homunculi; see \cite{knut:96} for discussion.)\footnote{``Computing by humans'' as a distinct practice I think would be self-defeating for both views, as if other things in nature also compute. Judging from pancomputationalism and ``nature computes'' enthusiasts, it seems difficult
to accept that the computer can be just a human concoction for thinking; see \cite{bozs:15a} for extensive discussion.}

The view I propose suggests representational and computational support
for harboring belief about past and future states and unknown events quite early on for humans,
first for planned action, giving representational value to past and future, then for language. Its consequences for level-1
    in working up to understanding a cultural belief system and its transmission across generations of level-3 subjects can be leveraged by the level-2 support in theory, rather than conflating the two levels, which is the preferred way to talk about human practice in neuroscience and dynamical approaches to mind. From the current perspective, one way to cope with change is to be responsive to the environment. Another way is to construct the environment or niche we choose to be in so that search for desirable inputs causes internal restructuring. Yet another is to control change centrally, in a way freezing it, for generations, as in rituals. All of this does something to change in human practice as we try to find ways to deal with it.

Mobile semantics, that is, representing, tracking and understanding change to support a basis for coping with it, from its abstraction to concrete realization, rather than  addressing spatiotemporality of change directly, is a  start for understanding the terms of serial or serializable high-level cognitive processes.
In doing so we cannot afford to alienate once again syntax from semantics, at least not from this kind of semantics.

\begin{spacing}{0}\setlength{\bibsep}{2pt}
\small
\bibliographystyle{LI-like}

\end{spacing}
\end{document}